\documentclass[a4wide]{article}
\usepackage[T1]{fontenc}

\usepackage{a4wide}    

\usepackage{graphicx}
\usepackage{amsmath}   
\usepackage{amsfonts}  
\usepackage{amssymb}   
\usepackage{adjustbox} 
\usepackage{multicol}  
\usepackage{booktabs}  

\usepackage[linesnumbered,ruled,vlined]{algorithm2e}

\SetCommentSty{mycommfont}

\SetKwInput{KwInput}{Input}                
\SetKwInput{KwOutput}{Output}              

\usepackage{float}     
\floatstyle{plaintop}  
\restylefloat{table}   

\usepackage{hyperref}
\hypersetup{
    colorlinks=true,
    linkcolor=blue,
    filecolor=magenta,      
    urlcolor=cyan,
}

\begin{document}

\title{ Improved Max-value Entropy Search for Multi-objective Bayesian Optimization with Constraints}

\date{}

\author{
  Daniel Fern\'andez-S\'anchez\\
  Universidad Aut\'onoma de Madrid\\
  Francisco Tom\'as y Valiente 11\\
  28049, Madrid, Spain\\
  \texttt{daniel.fernandezs@uam.es}
  \and
  Eduardo C. Garrido-Merch\'an\\
  Universidad Aut\'onoma de Madrid\\
  Francisco Tom\'as y Valiente 11\\
  28049, Madrid, Spain\\
  \texttt{eduardo.garrido@uam.es}
  \and
  Daniel Hern\'andez-Lobato\\
  Universidad Aut\'onoma de Madrid\\
  Francisco Tom\'as y Valiente 11\\
  28049, Madrid, Spain\\
  \texttt{daniel.hernandez@uam.es}
}

\maketitle              

\vspace{-.5cm}
\begin{abstract}
We present MESMOC+, an improved version of Max-value Entropy search for Multi-Objective
Bayesian optimization with Constraints (MESMOC). MESMOC+ can be used to solve constrained
multi-objective problems when the objectives and the constraints are expensive to evaluate. MESMOC+
works by minimizing the entropy of the solution of the optimization problem in function
space, i.e., the Pareto frontier, to guide the search for the optimum. The cost
of MESMOC+ is linear in the number of objectives and constraints. Furthermore, it is often
significantly smaller than the cost of alternative methods based on minimizing the entropy
of the Pareto set. The reason for this is that it is easier to approximate the required
computations in MESMOC+. Moreover, MESMOC+'s acquisition function is expressed as the sum of
one acquisition per each black-box (objective or constraint). Thus, it can be used in a decoupled
evaluation setting in which one chooses not only the next input location to evaluate, but
also which black-box to evaluate there. We compare MESMOC+ with related methods in synthetic
and real optimization problems. These experiments show that the entropy estimation provided
by MESMOC+ is more accurate than that of previous methods. This leads to better
optimization results. MESMOC+ is also competitive with other information-based methods
for constrained multi-objective Bayesian optimization, but it is significantly faster.
\end{abstract}

\section{Introduction}  \label{SEC:INTRO}

There are many problems where the purpose is to optimize several objectives 
$f_1(\mathbf{x}), \ldots, f_K(\mathbf{x})$ while fulfilling certain constraints 
$c_1(\mathbf{x}), \ldots, c_C(\mathbf{x})$, where $K$ and $C$ are the number of objectives and 
constraints. Also, normally, the input space $\mathcal{X}$ is bounded, \emph{i.e.} 
we optimize in $\mathcal{X} \subset \mathbb{R}^d$, where $d$ is the dimensionality of $\mathcal{X}$.
For example, one might want to maximize the speed of a robot 
while simultaneously minimizing its energy consumption \cite{ariizumi2014expensive}. 
Moreover one would like to avoid breaking any of its joints. To achieve this one could change 
the dimensions of the robot's gears and the materials of its manufacturing process. Another example would be to 
minimize the classification error of a deep neural network while at the same time 
minimizing the time needed to predict and not exceeding a certain amount of 
memory space. In this second example, one could modify the learning rate of the 
network, the number of hidden layers and the number of neurons in each hidden layer.

In problems where several objectives are optimized, most of the time there 
is no single optimal point, but a set of optimal points: the Pareto set 
$\mathcal{X}^\star$ \cite{collette2004multiobjective}. The objective values associated to the points 
in $\mathcal{X}^\star$ constitute the Pareto front $\mathcal{Y}^\star$. All 
the points in the Pareto set are optimal because they are not \emph{Pareto dominated} 
by any other point in $\mathcal{X}$. In a minimization context, a point $\mathbf{x}_1$ \emph{Pareto dominates}
$\mathbf{x}_2$ if $f_k(\mathbf{x}_1) \leq f_k(\mathbf{x}_2)$, 
$\forall k = \{1, ..., K\}$, with at least one strictly minor inequality. 
This means that it is not possible to improve the value in one 
objective without deteriorating the values obtained in others.  Moreover, the 
points of the Pareto set must be valid, \emph{i.e.}, they must satisfy all the 
constraints $c_j(\mathbf{x}) \geq 0$, $\forall j = \{1, ..., C\}$. On the other 
hand, as generally the potential size of $\mathcal{X}^\star$ is infinite (and 
therefore also that of $\mathcal{Y}^\star$), it is necessary to approximate the Pareto set.
  
The optimization problems described above have three main characteristics. 
First, there is no analytical form for the objectives or the constraints, 
thus they can be considered black-boxes. Second, the evaluations may be contaminated by noise. 
Third, new evaluations are quite expensive in some way, \emph{e.g.}, economically or temporally. 
In the example of the robot, we do not know beforehand what will be its speed and power 
consumption given some gear. Furthermore, building and testing the robot are likely to introduce some
noise into the results, as many external factors can influence these processes. Moreover, these processes
could be expensive economically and temporarily since we have to build the robot and prepare its tests.
To solve this type of problems while minimizing the evaluations performed
we can use a set of techniques called Bayesian optimization (BO) \cite{brochu2009tutorial}.

BO has two key pieces. First, a probabilistic model that estimates the potential values of the
black-boxes in unexplored regions of $\mathcal{X}$. The probabilistic model usually used is a
Gaussian process (GP). GPs are completely defined by a mean function $m(\cdot)$ and a covariance
function $v(\cdot,\cdot)$ \cite{rasmussen2006gaussian}. Second, an acquisition function that
measures the expected utility of evaluating at each point of X using the information provided
by the probabilistic model.
At each iteration of the BO algorithm, it evaluates the selected point, updates the probabilistic 
model and computes and maximizes the acquisition function. All this to find the point to evaluate in the next iteration. 
This process is repeated for a fixed number of iterations. Once the BO algorithm has finished,
the probabilistic models are optimized to estimate the solution of the problem. The acquisition
function is cheap to evaluate, unlike the black-boxes. Therefore, BO methods use the
probabilistic models to guide the search and save expensive evaluations \cite{shahriari2015taking}.

We have extended the acquisition function \emph{max-value entropy search} (MES) \cite{wang2017max}, 
which is based on the reduction of the entropy of the solution of the optimization problem in function space, 
to work with several objectives and constraints simultaneously. 
In the literature, there have already been some attempts to perform such an extension
\cite{belakaria2020max}. Nevertheless, the approximations of the acquisition function, which is
intractable, are very crude. In particular, the method proposed in \cite{belakaria2020max} simply
tries to maximize each objective and constraint independently. Our approximation of 
the exact acquisition is more accurate, as shown by our experiments, 
which leads to better optimization results. We call our method 
\emph{improved max-value entropy search for multi-objective optimization with constraints} (MESMOC+). As MES, MESMOC+ 
chooses as the next point to evaluate as the one at which the entropy of the Pareto front 
$\mathcal{Y}^\star$ is expected to be reduced the most. The reduction of entropy of $\mathcal{Y}^\star$ 
means that more information about the solution of the problem is available \cite{villemonteix2009informational,hennig2012entropy}. 
Several experiments involving synthetic and real optimization problems show that MESMOC+ outperforms
the method proposed in \cite{belakaria2020max}. Moreover, it obtains similar and sometimes even better 
results than those obtained by the best current acquisition functions for multi-objective optimization 
with several constraints from the literature \cite{garrido2019predictive}. Nevertheless, its 
computational cost per iteration is significantly smaller.
  
MESMOC+ is expressed as a sum of acquisitions, one per each black-box. Therefore, it can be used 
in a decoupled evaluation setting \cite{hernandez2015predictive}. More precisely, often the evaluation 
of black-boxes involves performing different experiments or simulations. In the example of the 
robot, we may perform a simulation to know if any joint breaks. Whereas to measure the speed or 
energy consumption it may be needed to manufacture and test the robot. 
Similarly, in the example of the neural network, knowing its classification 
error requires training and validation. By contrast, to determine its prediction time, or if 
it needs more memory than available, it is only necessary to build it, using, \emph{e.g.}, 
random weights. In a decoupled setting MESMOC+ chooses not only the next 
input location to evaluate, but also which black-box to evaluate. 
Our experiments compare the coupled and decoupled variants of MESMOC+ showing that sometimes
a decoupled evaluation setting gives better results over a coupled one in which all black-boxes
are evaluated at the same location.  

\vspace{-.25cm}
\section{Improved MES for Several Objectives and Constraints}  \label{SEC:MESMOC+}

In this section, we give the details of the proposed acquisition function
\emph{improved max-value entropy search for multi-objective optimization with constraints} (MESMOC+). 
In BO, the maximum of the acquisition function indicates the next point at which to evaluate the black-boxes.
For this, the information provided by the probabilistic models is used and each time that a new point is
evaluated, the probabilistic models are updated. Usually, Gaussian processes (GPs)
are the probabilistic models used \cite{shahriari2015taking}. Here we assume that the black-boxes 
are generated from a GP a priori with i.i.d. Gaussian noise with zero mean \cite{rasmussen2006gaussian}. 
For simplicity, the development of MESMOC+ is carried out considering a coupled evaluation setting, in which 
all black-boxes are evaluated at the same point. However, later on we explain how to use MESMOC+ in a decoupled setting.

Let $\mathcal{D} = \{(\mathbf{x}_n, \mathbf{y}_n)\}_{n=1}^N$ be the dataset with the evaluations 
performed up to iteration $N$, where $\mathbf{x}_n$ is the input evaluated in the $n$-th iteration 
and $\mathbf{y}_n$ is a vector with the values obtained when evaluating the $K$+$C$ black-boxes 
in $\mathbf{x}_n$, \emph{i.e.}, $\mathbf{y}_n = (f_1(\mathbf{x}_n), \ldots, f_K(\mathbf{x}_n), 
c_1(\mathbf{x}_n), \ldots, c_C(\mathbf{x}_n))$. Since MESMOC+ believes in reducing the entropy 
of the solution in the functional space after a $\mathbf{x}_{N+1}$ evaluation, the MESMOC+ acquisition 
function is:
\begin{align}
	\alpha(\mathbf{x}) &= H \left( \mathcal{Y}^\star | \mathcal{D} \right) -
	\mathbb{E}_{\mathbf{y}} \left[ H \left( \mathcal{Y}^\star| \mathcal{D} \cup \{(\mathbf{x}, \mathbf{y})\} \right)
	\right]\,, \label{EQ:MESMOC+INI1} 
\end{align}
where $H \left( \mathcal{Y}^\star| \mathcal{D} \right)$ is the entropy of the Pareto front $\mathcal{Y}^\star$ 
given by the probabilistic models, adjusted using the current dataset $\mathcal{D}$;  
the expectation is calculated over the potential values for $\mathbf{y}$ at $\mathbf{x}$, according to 
the GPs; and $H( \mathcal{Y}^\star| \mathcal{D} \cup \{(\mathbf{x}, \mathbf{y})\} )$ is the entropy 
of $\mathcal{Y}^\star$ after including the new data point $(\mathbf{x}, \mathbf{y})$ in the dataset.

Critically, evaluating the entropy of $\mathcal{Y}^\star$ is very challenging. 
In order to avoid this problem, we can rewrite \eqref{EQ:MESMOC+INI1} in an equivalent
form, as suggested in \cite{wang2017max}, by noting that \eqref{EQ:MESMOC+INI1} is exactly 
the mutual information between $\mathcal{Y}^\star$ and $\mathbf{y}$ \cite{hernandez2014predictive,hernandez2016predictive}. 
Therefore, since $I(\mathcal{Y}^\star; \mathbf{y}) = I(\mathbf{y}; \mathcal{Y}^\star)$, then, we
can swap the roles of $\mathcal{Y}^\star$ and $\mathbf{y}$ in (\ref{EQ:MESMOC+INI1}) and the MESMOC+ acquisition
function is:
\begin{align}
	\alpha(\mathbf{x}) &= H \left( \mathbf{y}| \mathcal{D}, \mathbf{x} \right) -
	\mathbb{E}_{\mathcal{Y}^\star} \left[ H \left( \mathbf{y}| \mathcal{D}, \mathbf{x}, \mathcal{Y}^\star \right) \right]\,,
	\label{EQ:MESMOC+INI2}
\end{align}
where $H \left( \mathbf{y}| \mathcal{D}, \mathbf{x} \right)$ is the entropy of 
$p\left( \mathbf{y}| \mathcal{D}, \mathbf{x} \right)$, \emph{i.e.}, the entropy of the predictive distribution
of the GPs at $\mathbf{x}$; now the expectation is with respect to potential values of 
$\mathcal{Y}^\star$; and $H \left( \mathbf{y}| \mathcal{D}, \mathbf{x}, \mathcal{Y}^\star \right)$ is the entropy 
of the predictive distribution conditioned to the Pareto front $\mathcal{Y}^\star$ being the solution of the problem.

The expression given in Eq. \eqref{EQ:MESMOC+INI2} is the acquisition function targeted by MESMOC+. 
Thus, in each iteration, the next query will be chosen at the maximum point \eqref{EQ:MESMOC+INI2}, 
\emph{i.e.} $\smash{\mathbf{x}_{N+1} = \arg \max_{\mathbf{x} \in \mathcal{X}} \alpha(\mathbf{x})}$. 
It is easier to work with this expression than with \eqref{EQ:MESMOC+INI1} because here we do 
not have to evaluate the entropy of $\mathcal{Y}^\star$, a set of probably infinite size. 
The first term of the r.h.s of \eqref{EQ:MESMOC+INI2} is simply the entropy of the current predictive 
distribution, and since we assume that there is no correlation between the black-boxes, 
its expression is the sum of the entropy of $K+C$ Gaussian distributions. Namely
\begin{align}\label{EQ:FISRTTERMMESMOC+}
	H \left( \mathbf{y}| \mathcal{D}, \mathbf{x} \right) &=
	\sum_{k=1}^K \frac{\log(2 \pi e v_k^f)}{2}  + \sum_{j=1}^C \frac{\log(2 \pi e v_j^c)}{2}\,,
\end{align}
where $v_k^f = v_k^f(\mathbf{x})$ and $v_j^c = v_j^c(\mathbf{x})$ are the predictive 
variance for the $k$-th and $j$-th objective and constraint respectively.
Nevertheless, the evaluation of the second term in the r.h.s. of \eqref{EQ:MESMOC+INI2} is intractable. 
The expectation can be approximated by generating Monte Carlo samples of $\mathcal{Y}^\star$, 
calculating the entropy of $p( \mathbf{y}| \mathcal{D}, \mathbf{x})$ and averaging the results. 
To generate samples of $\mathcal{Y}^\star$, we first use a random feature approximation 
of the GPs (see \cite{rahimi2007random} for further details) to generate samples of the 
objectives and constraints. These samples 
are then optimized using a grid of points, as in \cite{garrido2019predictive}, to sample $\mathcal{Y}^\star$. 
This step is cheap because, unlike the actual black-boxes, we can evaluate the GP samples with 
small cost. Other GP sampling methods may be used instead, \emph{e.g.}, 
\cite{wilson2020efficiently}. Finally, on the other hand, the evaluation 
of the entropy of $p(\mathbf{y}| \mathcal{D}, \mathbf{x}, \mathcal{Y}^\star)$ has to 
be approximated. We explain the approximation employed in the next section.

\subsection{Approximating the Conditional Predictive Distribution}  \label{SB:CPDADF}
  
As in the original formation of \emph{max-value entropy search} (MES) \cite{wang2017max},
we consider that the evaluations are noiseless.  Namely, we approximate 
$p(\mathbf{f}, \mathbf{c}| \mathcal{D}, \mathbf{x}, \mathcal{Y}^\star)$ instead of 
$p(\mathbf{y}| \mathcal{D}, \mathbf{x}, \mathcal{Y}^\star)$, where 
$\mathbf{f}=\{f_1(\mathbf{x}), \ldots, f_K(\mathbf{x})\}$ and
$\mathbf{c}=\{c_1(\mathbf{x}), \ldots, c_C(\mathbf{x})\}$ 
are the predicted values for the black-boxes. At the end of the next section,
we modify the acquisition function developed to take additive noise into account. 
The expression of $p(\mathbf{f}, \mathbf{c}| \mathcal{D}, \mathbf{x}, \mathcal{Y}^\star)$ is 
obtained using Bayes' rule. Namely,
\begin{align}
	p(\mathbf{f}, \mathbf{c}| \mathcal{D}, \mathbf{x}, \mathcal{Y}^\star)
	& = Z^{-1} p(\mathbf{f}, \mathbf{c}| \mathcal{D}, \mathbf{x})
	p(\mathcal{Y}^\star| \mathbf{f}, \mathbf{c})\,,
\end{align}
where $Z^{-1}$ is a normalization constant, $p(\mathbf{f}, \mathbf{c}| \mathcal{D}, \mathbf{x})$ 
is the probability of the objectives and constraints given $\mathcal{D}$ and $\mathbf{x}$, 
and $p(\mathcal{Y}^\star| \mathbf{f}, \mathbf{c})$ is the probability that $\mathcal{Y}^\star$ 
is a valid Pareto front given $\mathbf{f}$ and $\mathbf{c}$.

The factor $p(\mathcal{Y}^\star| \mathbf{f}, \mathbf{c})$ in \eqref{EQ:CPDM1} removes all 
configurations of the objectives and constraints values, $(\mathbf{f}, \mathbf{c})$, that 
are incompatible with $\mathcal{Y}^\star$ being the Pareto front of the problem.
Therefore, $p(\mathcal{Y}^\star| \mathbf{f}, \mathbf{c})$ must be $0$ when 
$\mathbf{c}$ is valid ($\mathbf{c}$ does satisfy $c_j (\mathbf{x}) \geq 0, \forall j \in \{1, ..., C\}$), 
but $\mathbf{f}$ is not \emph{Pareto dominated} by all the points in the Pareto front, \emph{i.e}, 
all $\mathbf{f}^\star \in \mathcal{Y}^\star$.  
Similarly, $p(\mathcal{Y}^\star| \mathbf{f}, \mathbf{c})$ will be $1$ if all points $\mathbf{f}^\star$ in the 
Pareto front $\mathcal{Y}^\star$ dominate $\mathbf{f}$, or if $\mathbf{c}$ is invalid (\emph{i.e.}, at least 
one constraint is negative at $\mathbf{x}$). This can be expressed, informally, as follows:
\begin{align}\label{EQ:PYSTAR}
  p(\mathcal{Y}^\star| \mathbf{f}, \mathbf{c})
  & \propto
	\prod_{\mathbf{f}^\star \in \mathcal{Y}^\star} 
	\big( 1 - \prod_{j=0}^C \Theta (c_j) \prod_{k=0}^K \Theta \left(f_k^\star  - f_k \right) \big) 
  \propto \prod_{\mathbf{f}^\star \in \mathcal{Y}^\star} \Omega (\mathbf{f}^\star, \mathbf{f}, \mathbf{c})
	\,,
\end{align}
where $\Theta(\cdot)$ is the Heaviside step function,
$f_k = f_k (\mathbf{x})$, $c_j = c_j (\mathbf{x})$, $f_k^\star$ is the $k$-th value of $\mathbf{f}^\star$ 
and $\smash{\Omega (\mathbf{f}^\star, \mathbf{f}, \mathbf{c}) = 1 - \prod_{j=0}^C \Theta (c_j (\mathbf{x})) 
\prod_{k=0}^K \Theta \left(f_k^\star - f_k (\mathbf{x}) \right)}$. 
Note that the value of \eqref{EQ:PYSTAR} will only be $1$, 
if $\Omega (\mathbf{f}^\star, \mathbf{f}, \mathbf{c})$ is $1$ for all the 
$\mathbf{f}^\star$ in $\mathcal{Y}^\star$. 
To make $\Omega (\mathbf{f}^\star, \mathbf{f}, \mathbf{c})$ be $1$, 
$\smash{ \prod_{j=0}^C \Theta (c_j (\mathbf{x}))}$ or
$\smash{\prod_{k=0}^K \Theta \left(f_k^\star - f_k (\mathbf{x}) \right)}$ have to be $0$, 
and that happens if all the values of $\mathbf{c}$ are greater or equal to $0$, or if all the 
values of $\mathbf{f}^\star$ are lower or equal to those of $\mathbf{f}$, except one which must 
be strictly minor. These are precisely the conditions described above Eq. (\ref{EQ:PYSTAR}).

Computing the normalization constant and the entropy of 
\eqref{EQ:CPDM1} is intractable. Thus, we must approximate 
this distribution. Critically, the approximation should be cheap.
For this, we use Assumed Density Filtering (ADF) \cite{boyen1998tractable,minka2001expectation}.
ADF simply approximates each non-Gaussian factor in \eqref{EQ:CPDM1} using a Gaussian distribution.
Since the predictive distribution of a GPs is Gaussian, the only non-Gaussian factors are the 
$\smash{\Omega (\mathbf{f}^\star, \mathbf{f}, \mathbf{c})}$ factors in \eqref{EQ:PYSTAR}.
We assume independence among the objectives and constraints, this results in a factorizing Gaussian
approximation of each factor. The specific updates for the parameters of each of these Gaussians 
are described in the supplementary material.  Because the Gaussian distribution is closed under the 
product operation, the approximation of \eqref{EQ:CPDM1} is a factorizing Gaussian distribution. 
Let $\mathbf{\tilde{m}}^{f}$ and $\mathbf{\tilde{m}}^{c}$ and the $\mathbf{\tilde{v}}^{f}$ and $\mathbf{\tilde{v}}^{c}$   
be respectively the means and variances of that approximation.

\subsection{The MESMOC+ Acquisition Function}  \label{SB:MESMOC+FINALEXPRESION}
   
After the execution of ADF has finished, the variances of the objectives and the 
constraints of the predictive distribution at the candidate point $\mathbf{x}$ 
conditioned to the Pareto front $\mathcal{Y}^\star$ are available. 
Therefore, to obtain the approximate expression of \eqref{EQ:MESMOC+INI2} one simply 
has to combine \eqref{EQ:FISRTTERMMESMOC+} with the result of the calculation of the 
entropy of $p(\mathbf{f}, \mathbf{c}| \mathcal{D}, \mathbf{x}, \mathcal{Y}^\star)$. 
Because this distribution is approximated with a Gaussian distribution using ADF, the 
approximate entropy has a form similar to that of \eqref{EQ:FISRTTERMMESMOC+}.
The consequence is that the acquisition function can be approximated simply as the
difference between the entropy of two factorizing multi-variate Gaussians. One for
the objectives and one for the constraints. Namely,
\begin{align} \label{EQ:MESMOC+NONOISY}
	\alpha(\mathbf{x})
	\approx &\sum_{k=1}^K\log(v_k^f) + \sum_{j=1}^C\log(v_j^c) 
	- \frac{1}{M} \sum_{m=1}^M \big[ \sum_{k=1}^K\log(\tilde{v}_k^f) + \sum_{j=1}^C\log(\tilde{v}_j^c)
	\big]\,,
\end{align}
where $M$ is the number of Monte Carlo samples of $\mathcal{Y}^\star$, 
$v_k^f = v_k^f(\mathbf{x})$, $v_j^c = v_j^c(\mathbf{x})$, 
$\tilde{v}_k^f = \tilde{v}_k^f(\mathbf{x}|\mathcal{Y}^\star_{(m)})$, $\tilde{v}_j^c = \tilde{v}_j^c(\mathbf{x}|\mathcal{Y}^\star_{(m)})$, 
are the approximate variances of the conditional distribution, 
and $\{\mathcal{Y}^\star_{(m)}\}^M_{m=1}$ is the set of Monte Carlo samples 
of $\mathcal{Y}^\star$. 

In order to take into account the noise of each black-box, one simply needs to add 
its variance to the variance of the corresponding objectives and constraints.
Unfortunately, the behavior of MESMOC+ when using \eqref{EQ:MESMOC+NONOISY} to approximate the acquisition function is 
not the expected one. More precisely, \eqref{EQ:MESMOC+NONOISY} is highly influenced by a small decrease in the variance of 
the conditional predictive distribution. This is particularly the case for points that have a very small associated initial 
variance, \emph{e.g.}, $10^{-5}$. The logarithm tends to amplify these small differences (\emph{e.g.}, a variance reduction 
from $10^{-5}$ to $10^{-6}$ will result in a log difference that is approximately equal to 2.32) 
and the consequence is a highly exploitative behavior of the BO method which tends to perform evaluations 
that are very close to points that have already been evaluated. To avoid this, we modified MESMOC+'s acquisition function
to take into account the absolute reduction in the variance instead. The final expression of \textrm{MESMOC+} acquisition is:
\begin{align}\label{EQ:MESMOC+FINAL}
  \alpha(\mathbf{x})
  \approx &\sum_{k=1}^K \left( v_k^f  + (\sigma_k^f)^2 \right)
  + \sum_{j=1}^C       \left( v_j^c  + (\sigma_j^c)^2 \right) \notag \\
	&- \frac{1}{M} \sum_{m=1}^M 
	\big[ \sum_{k=1}^K \left( \tilde{v}_k^f + (\sigma_k^f)^2 \right) +   
	\sum_{j=1}^C \left(\tilde{v}_j^c + (\sigma_j^c)^2 \right)
  \big]\,,
\end{align}
where $(\sigma_k^f)^2$ and $(\sigma_j^c)^2$ are the noise variances of each objective and constraint.
  
Note that \eqref{EQ:MESMOC+FINAL} is a sum of one acquisition per black-box. 
Namely, $\alpha(\mathbf{x})= \sum_{k=1}^K \alpha_k^f(\mathbf{x}) + \sum_{j=1}^C \alpha_j^c(\mathbf{x})$. Therefore,
\eqref{EQ:MESMOC+FINAL} can be readily used in a decoupled evaluation setting. In this case, when all black-boxes 
are competing to be evaluated, each individual acquisition function is maximized separately. The black-box with the 
maximum value associated to the acquisition is chosen for evaluation.
  
The cost evaluating \eqref{EQ:MESMOC+FINAL} is in $\mathcal{O}(\sum_{m=1}^M (K+C)|\mathcal{Y}_{(m)}^\star|)$, where $M$ is the number of 
Monte Carlo samples, and $K$ and $C$ are the number of objectives and constraints respectively. The part of the 
cost corresponding to $(K+C)|\mathcal{Y}_{(m)}^\star|$ comes from running the ADF algorithm to approximate 
the variances of predictive distribution conditioned to $\mathcal{Y}^\star$. This approximation is run for each candidate 
point $\mathbf{x}$ at which the acquisition needs to be evaluated. For each sample of the objectives and constraints, 
$\mathcal{Y}^\star$ is approximated using $50$ points. The acquisition function is optimized 
using a Quasi-Newton method with the gradient approximated by differences. A grid is used to find a good starting value.

\section{Related Work} \label{SEC:RELATEDWORK}

There are other acquisition functions that can deal with multiple objectives and constraints. 
They are described in this section and compared to MESMOC+.  

Bayesian Multi-objective optimization (BMOO) extends the Pareto dominance rule 
to introduce a preference to perform evaluations at points that are more 
likely to be feasible \cite{feliot2017bayesian}. This extended rule comes from the fact that in constrained problems there may be no 
feasible point observed. The extended rule simply applies a transformation to the two points that are 
compared to see if one dominates the other. This transformation function is:
\begin{align}\label{EQ:BMOOPARETORULE}
	\Psi(\mathbf{y}^f, \mathbf{y}^c) =
	\left\{
	\begin{matrix}
	(\mathbf{y}^f, \mathbf{0}) \qquad &\text{if} \ \mathbf{y}^c \geq 0
	\\
	(+\infty, \min(\mathbf{y}^c, \mathbf{0})) \ &\text{otherwise}
	\end{matrix}
	\right.\,,
\end{align}
where $\mathbf{y}^f$ and $\mathbf{y}^c$ are the vectors observations of the objectives and 
constraints, respectively.  BMOO uses the acquisition function expected improvement where 
improvement is measured in terms of hyper-volume. The hyper-volume is simply the volume of points 
in functional space above the best observed points using the extended Pareto dominance rule.
It is maximized by the actual solution of the problem. The acquisition function of BMOO measures the 
expected hyper-volume improvement in the extended space:
\begin{equation}\label{EQ:BMOO}
	\alpha(\mathbf{x}) = \mathbb{E}_{\mathbf{y}^f, \mathbf{y}^c} \left [
	\int_{\mathcal{G}_N} \mathbb{I}(\Psi(\mathbf{y}^f, \mathbf{y}^c) \prec \Psi(\mathbf{y}) ) d\mathbf{y}
	\right]
\end{equation}
where $\mathbf{a} \prec \mathbf{b}$ means that $\mathbf{a}$ is Pareto dominated 
by $\mathbf{b}$, $\mathbb{I}(\cdot)$ is the indicator function, $\mathcal{G}_N$ is 
the set of points not dominated until iteration $N$, and the expectation is w.r.t the 
predictive distribution of the GPs.  Since \eqref{EQ:BMOO} cannot be calculated analytically,
in \cite{feliot2017bayesian} the authors decided to
swap the expectation and  the integral, and to approximate the integral by using Monte Carlo
samples from a uniform distribution in $\mathcal{G}_N$. However, generating these samples
is very expensive. A Metropolis-Hastings algorithm is suggested for this. BMOO was initially 
described for noiseless scenarios, but the method can also be applied when the black-boxes 
are contaminated with noise \cite{garrido2019predictive}. Finally, BMOO is often outperformed 
by another acquisition function known as PESMOC and it does not allow for decoupled evaluations \cite{garrido2019predictive}.

PESMOC is another acquisition function that  focuses on the reduction of the entropy of the 
solution of the optimization problem \cite{garrido2019predictive}. However, it targets 
the entropy of the Pareto set $\mathcal{X}^\star$, instead of the entropy of the Pareto frontier $\mathcal{Y}^\star$. 
The acquisition function is hence the expected reduction in the entropy of $\mathcal{X}^\star$.  As in MESMOC+, the entropy of 
$\mathcal{X}^\star$ is intractable and requires complicated approximations. PESMOC 
rewrites the expression of the acquisition function noting that the expected reduction in the entropy of 
$\mathcal{X}^\star$ is the mutual information between $\mathcal{X}^\star$ and $\mathbf{y}$. Because
the mutual information is symmetric, it is equivalent to the mutual information between 
$\mathbf{y}$ and $\mathcal{X}^\star$. Doing this rewriting, the acquisition function of PESMOC is:
\begin{equation}\label{EQ:PESMOC}
	\alpha(\mathbf{x})
	= H \left( \mathbf{y}| \mathcal{D}, \mathbf{x} \right) - \mathbb{E}_{\mathcal{X}^\star} \left[ H \left(
			\mathbf{y}| \mathcal{D}, \mathbf{x}, \mathcal{X}^\star \right) \right]\,,
\end{equation}
where the first term of the r.h.s. is the same as in MESMOC+. Namely, the entropy of the 
predictive distribution. The expectation is w.r.t. the Pareto set $\mathcal{X}^\star$ instead of 
$\mathcal{Y}^\star$. Finally, the second term of the r.h.s. is the entropy of the predictive 
distribution conditioned to $\mathcal{X}^\star$ being the solution to the problem.

As in MESMOC+, the second term of the r.h.s. of \eqref{EQ:PESMOC} is intractable 
and must be approximated. The expectation is approximated also by a Monte Carlo average, as in MESMOC+.
The method for generating the  samples of $\mathcal{X}^\star$ is equivalent to the one used in MESMOC+.
The entropy of $p(\mathbf{y}|\mathcal{D}, \mathbf{x}, \mathcal{X}^\star)$ needs to be approximated. 
Expectation propagation is used for that purpose \cite{minka2001expectation}.
However, and importantly, this step is more complicated than in MESMOC+, where the entropy of 
$p(\mathbf{y}|\mathcal{D}, \mathbf{x}, \mathcal{Y}^\star)$ has to be approximated instead. 
In particular, there are more non-Gaussian factors that need to be approximated in PESMOC, and 
the approximation is more complicated since some of the factors depend on two variables, which 
involves working with bi-variate Gaussians. By contrast, all the factors 
in MESMOC+ are univariate which means that only one-dimensional Gaussians have to be 
used in practice.
This is results in MESMOC+ acquisition being significantly less expensive to compute and easier to implement.
Our experiments also show that MESMOC+ gives similar results to those of PESMOC.

Max-value entropy search (MES) has also used to address optimization problems that involve 
a single objective and no constraints \cite{wang2017max}, a single objective and 
several constraints \cite{perrone2019constrained} and with several objectives and no 
constraints \cite{belakaria2019max,suzuki2020multi}. Notwithstanding, none of these 
methods can address several objectives and constraints at the same time.
Moreover, the extension to several constraints or several objectives is not trivial at all.

MESMOC is an acquisition function developed in an independent work 
\cite{belakaria2020max}, which also minimizes the entropy of $\mathcal{Y}^\star$, as MESMOC+ does.
The expression considered by MESMOC for the acquisition function is also 
\eqref{EQ:MESMOC+INI2}.  However, the proposed approximation for the entropy 
of $p(\mathbf{y}|\mathcal{D}, \mathbf{x}, \mathcal{Y}^\star)$ is different.
Instead of minimizing the objectives, in \cite{belakaria2020max} maximization is considered.
Ignore the constraints initially. 
Let the sampled Pareto frontier be $\mathcal{Y}^\star=\{\mathbf{z}_1,\ldots,\mathbf{z}_m\}$ with $m$ 
the size of $\mathcal{Y}^\star$. In \cite{belakaria2020max} 
they argue that a sufficient condition for some point $\mathbf{y}$ being 
compatible with $\mathcal{Y}^\star$ as the solution of the problem is that
$y^j \leq \max \{z_1^j,\ldots,z_m^j\} \ \forall j \in \{1,\ldots,K\}$.
That is, the value of $\mathbf{y}$ for the $j$-th objective cannot be better 
than the maximum value for that objective, according to $\mathcal{Y}^\star$.
However, this condition is not complete because $\mathbf{y}$ can be optimal, (\emph{i.e.}, $\mathbf{y}$ is
incompatible with $\mathcal{Y}^\star$) even if none of its values are greater than 
the maximum value for the corresponding objective. E.g., let $K=2$ and $\mathcal{Y}^\star = \{(1,0), (0,1)\}$. Consider now the point
$(0.7, 0.7)$. Their components are lower than $1=\max\{z_1^j,\ldots,z_m^j\} \forall j \in \{1,\ldots,K\}$, but 
this point is optimal and non-dominated by any of the points in $\mathcal{Y}^\star$.
Then, the constraints are incorporated in \cite{belakaria2020max}, in an ad-hoc way, simply by enforcing that 
$c_j(\mathbf{x}) \leq \text{max}\{\tilde{z}_1^j,\ldots,\tilde{z}_m^j\}$ for 
$j=1,\ldots,C$, where $\{\tilde{\mathbf{z}}_i\}_{i=1}^m$ are the constraint values associated 
to the points in $\mathcal{Y}^\star$. That is, the constraint values have to be 
smaller than the maximum constraint values associated to the Pareto front $\mathcal{Y}^\star$,
as it was done with the objectives. Notwithstanding, in the provided code in \cite{belakaria2020max}, 
the implementation considered not only $\mathcal{Y}^\star$, but all the evaluations performed so far. 
The consequence is that the acquisition proposed in \cite{belakaria2020max} is simply
the sum of the MES acquisition function for each objective and constraint \cite{wang2017max}.
This makes sense, and is equivalent to maximizing all the objectives and constraints independently.
Maximizing the objectives is expected to give good solutions. Maximizing the constraints is expected
to provide feasible solutions. 

The optimization of the resulting acquisition function is, however,
restricted in \cite{belakaria2020max} to those regions of the input space in which 
the GP means for the constraints are strictly positive. This becomes problematic in 
problems in which finding feasible points is difficult. In particular, if all the observations 
are infeasible, the GP means for the constraints will be negative in all the input space 
(even though the associated GP variance can be high). In that case, we simply choose at random 
the next point to evaluate.

Our experiments show that the accuracy of MESMOC for approximating the acquisition in Eq. (\ref{EQ:MESMOC+INI2})
is worse than that of our method, MESMOC+. Furthermore, in several optimization problems MESMOC+ outperforms MESMOC.
We believe this is related to the accuracy of the approximation and the problem of MESMOC for finding feasible
solutions.

\section{Experiments}  \label{SEC:EXPERIMENTS}

We compare MESMOC+ and its decoupled variant MESMOC+$_\text{dec}$ with the 
acquisition functions described in Section \ref{SEC:RELATEDWORK} (\emph{i.e.}, BMOO, 
PESMOC, and MESMOC) and with a random search (RANDOM). 
BMOO and PESMOC are provided in the Bayesian optimization software 
Spearmint (\url{https://github.com/EduardoGarrido90/Spearmint}). We have also implemented in that software 
MESMOC+ and MESMOC, closely following the code provided in \cite{belakaria2020max}. See the supplementary 
material. We use a Mat\'ern52 with ARD as the kernel of all GPs, and to learn their hyper-parameters we 
use slice sampling with 10 samples, as does in \cite{snoek2012practical}. This is also the number of samples 
considered in MESMOC+, MESMOC and PESMOC for $\mathcal{Y}^\star$, $\mathcal{Y}^\star$ and $\mathcal{X}^\star$, 
respectively. To maximize the acquisition function we use L-BFGS using a grid of 1.000 points to 
choose the starting position. The gradients of the acquisition function are approximated by differences. 
All the experiments are repeated 100 times and we report average results. The recommendation 
of each method is obtained by optimizing the means of the GPs at each iteration. 
We follow the approach suggested in \cite{garrido2019predictive} to avoid recommending infeasible solutions. 

\subsection{Quality of the Approximation of the Acquisition Function}  \label{SB:QUALITYAPPROXIMATION}

We compare in a simple problem the acquisition function of MESMOC+ and MESMOC 
with the exact acquisition function described in Eq. (\ref{EQ:MESMOC+INI2}).
The problem considered has only two objectives and one constraint. In this setting, 
quadrature methods are feasible to evaluate the entropy of
$p(\mathbf{y}| \mathcal{D}, \mathbf{x}, \mathcal{Y}^\star)$ at a much higher computational cost.
They are expected to provide an approximation that is almost equal to that of the exact acquisition.

\begin{figure*}[tbh]\label{FIG:CMPACQ}
	\centering
	\resizebox{\textwidth}{!}{\begin{tabular}{cc}
		\includegraphics[width=0.5\textwidth]{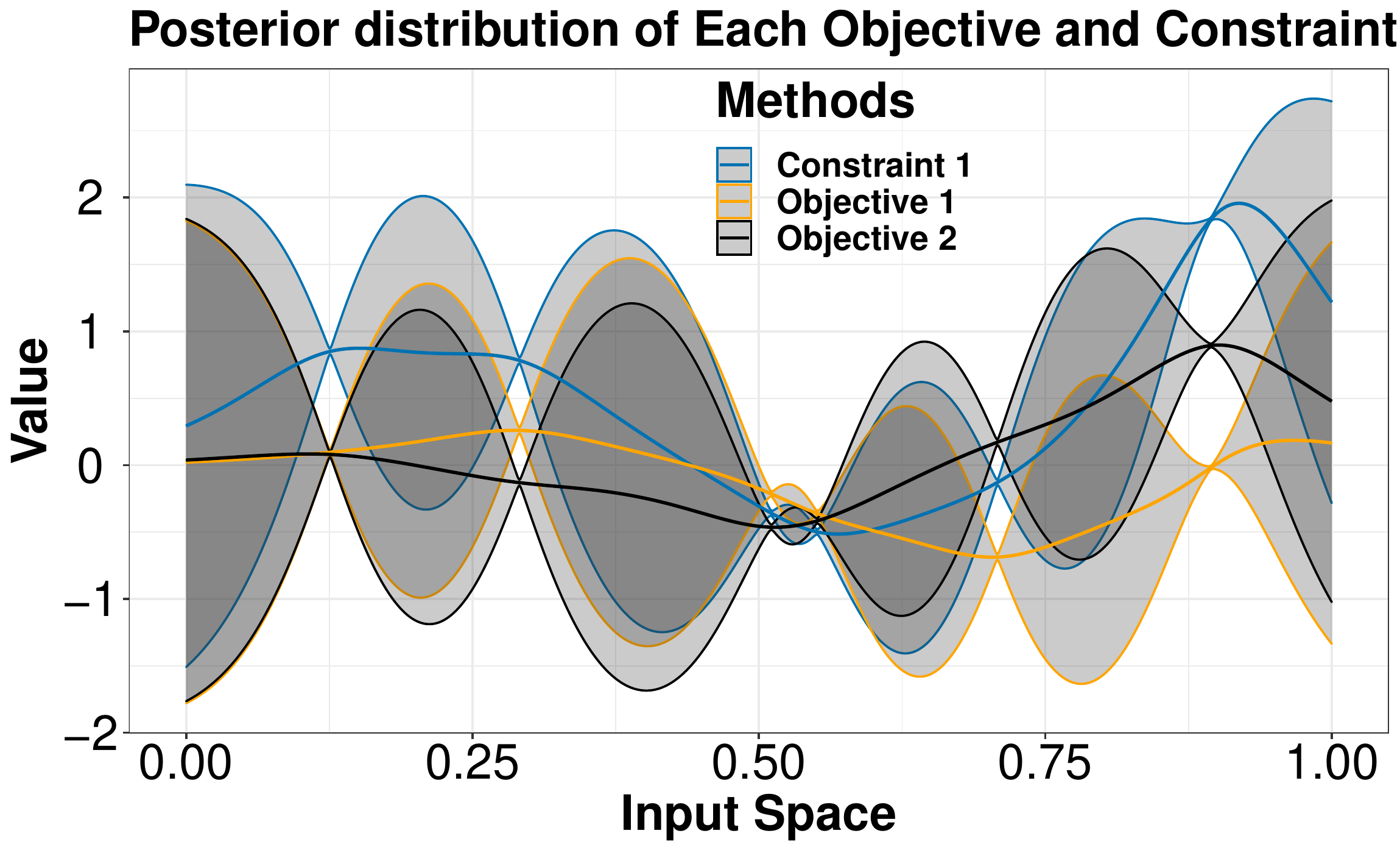}
		\includegraphics[width=0.5\textwidth]{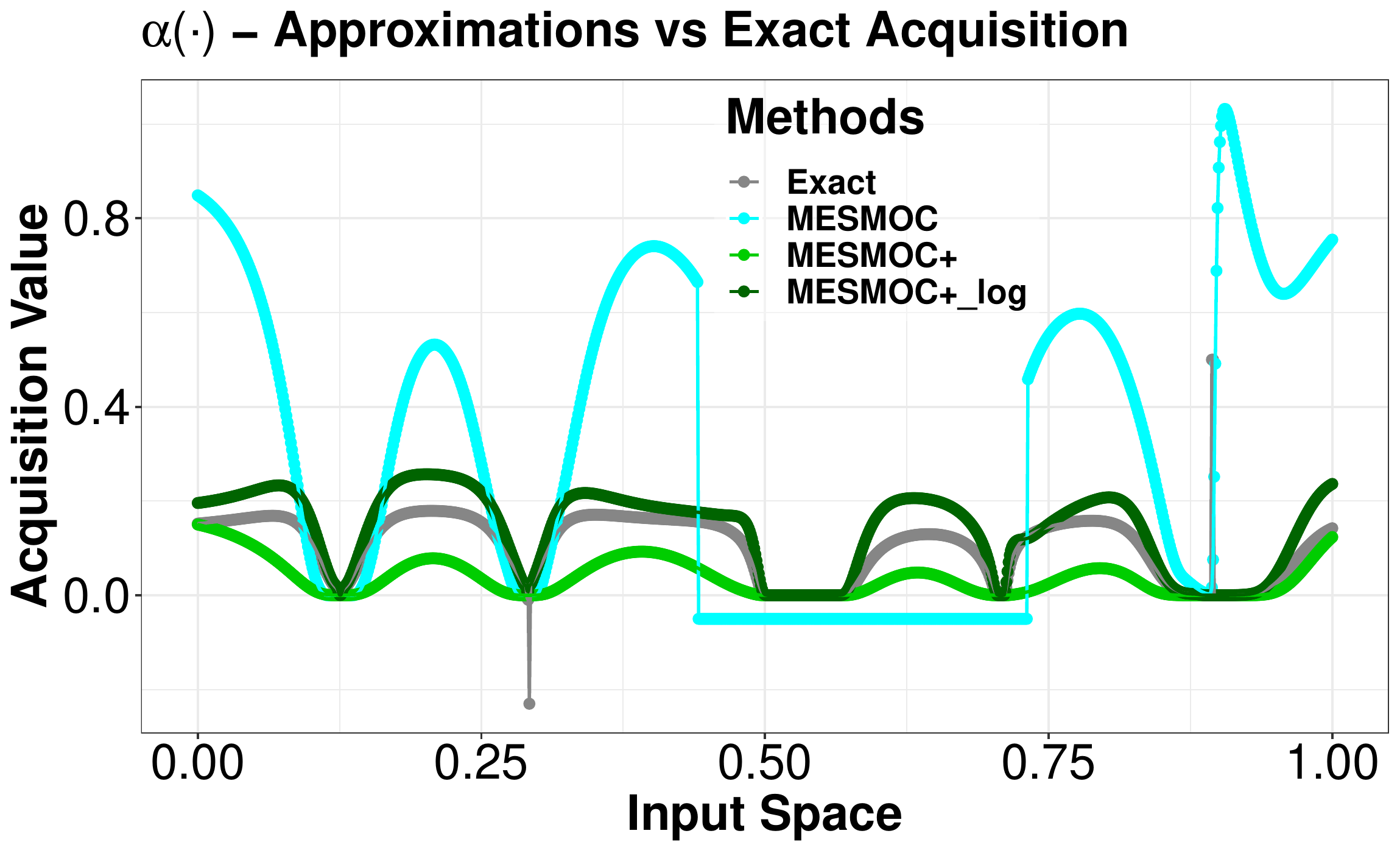}
	\end{tabular}}
	\caption{(left) GP predictive distributions for the objectives and constraints. 
		(right) The corresponding estimated acquisition function of each method, MESMOC+ and MESMOC, 
		and the exact acquisition (Exact). Best seen in color. }
\end{figure*}

The left column of the Fig. \ref{FIG:CMPACQ} shows the current observations and predictive 
distributions for the objectives and constraints. The right column shows the sum of the acquisition 
function for MESMOC and MESMOC+. In the case of MESMOC+, we show results for the proposed method and 
when the log of the variance is considered (MESMOC+$_\text{log}$). See Eq. \eqref{EQ:MESMOC+NONOISY}. 
Last, we also show the results of the quadrature method (Exact). We note that the 
approximation of MESMOC+$_\text{log}$ seems to be the most accurate, closely followed 
by MESMOC+. By contrast, the approximation of MESMOC does not look similar to the exact acquisition. 
MESMOC avoids evaluations in the region where the GP mean of the constraint is negative.
MESMOC's acquisition there correspond to a constant value smaller than zero.
A comparison of the quality of the approximation in a decoupled scenario is found in the supplementary material. 
There, MESMOC+ also provides a more accurate approximation than MESMOC.

\subsection{Synthetic Experiments}  \label{SB:SYTHETICEXP}

Here, the objectives and the constraints are sampled from a GP. 
We also consider two scenarios. One with noiseless observations and another where 
the observations are contaminated with standard Gaussian noise with variance $0.1$.
The first experiment has 4 dimensions, 2 objectives and 2 constraints, and the second has 6 dimensions, 
4 objectives and 2 constraints. The performance of each method is measured as the relative difference 
(in log-scale) of the hyper-volume of the recommendation made and the maximum 
hyper-volume, with respect to the number of evaluations made. 

\begin{figure*}[tbh]\label{FIG:EXPSYNTHE}
	\centering
	\resizebox{\textwidth}{!}{\begin{tabular}{cc}
		\includegraphics[width=0.5\textwidth]{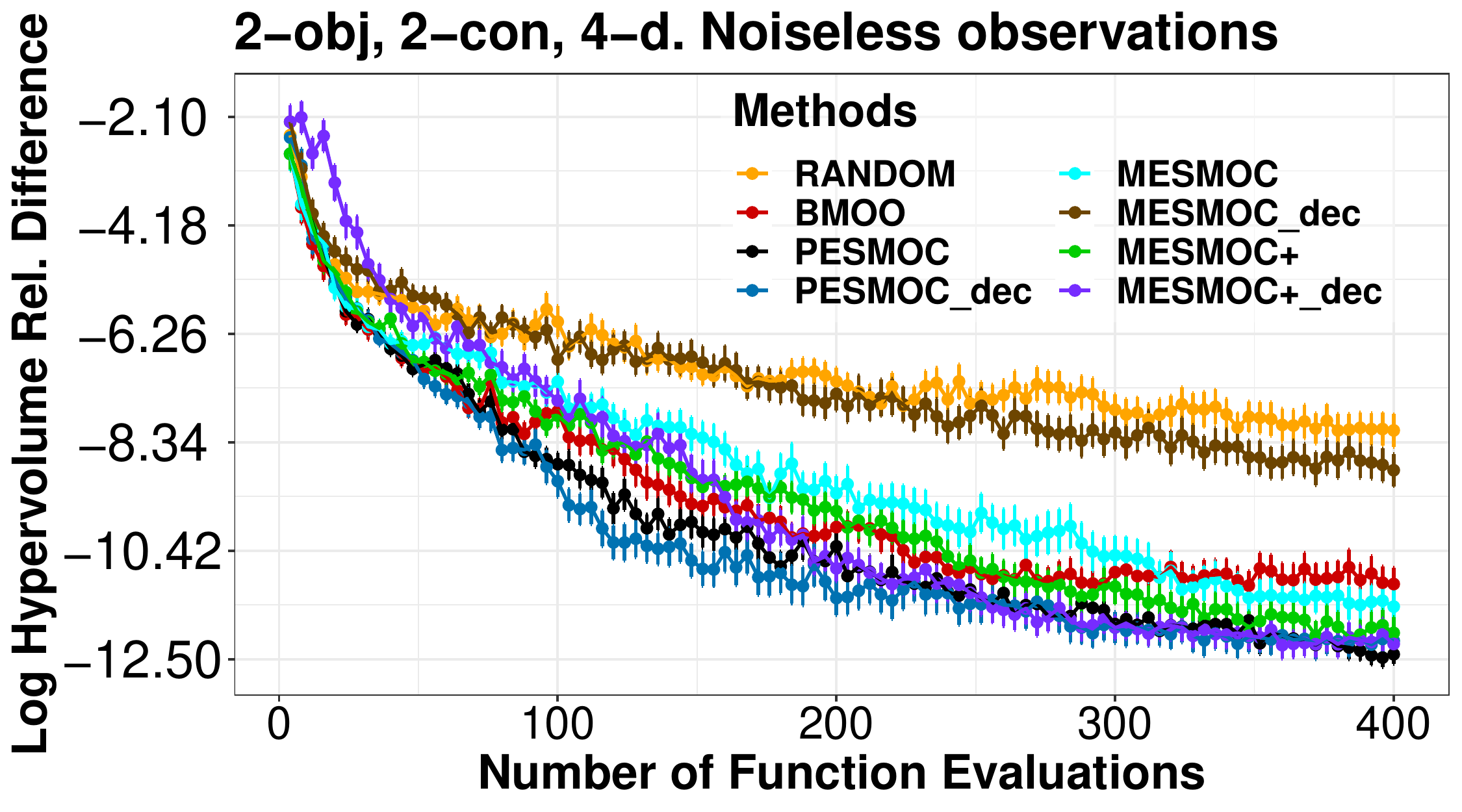}
		\includegraphics[width=0.5\textwidth]{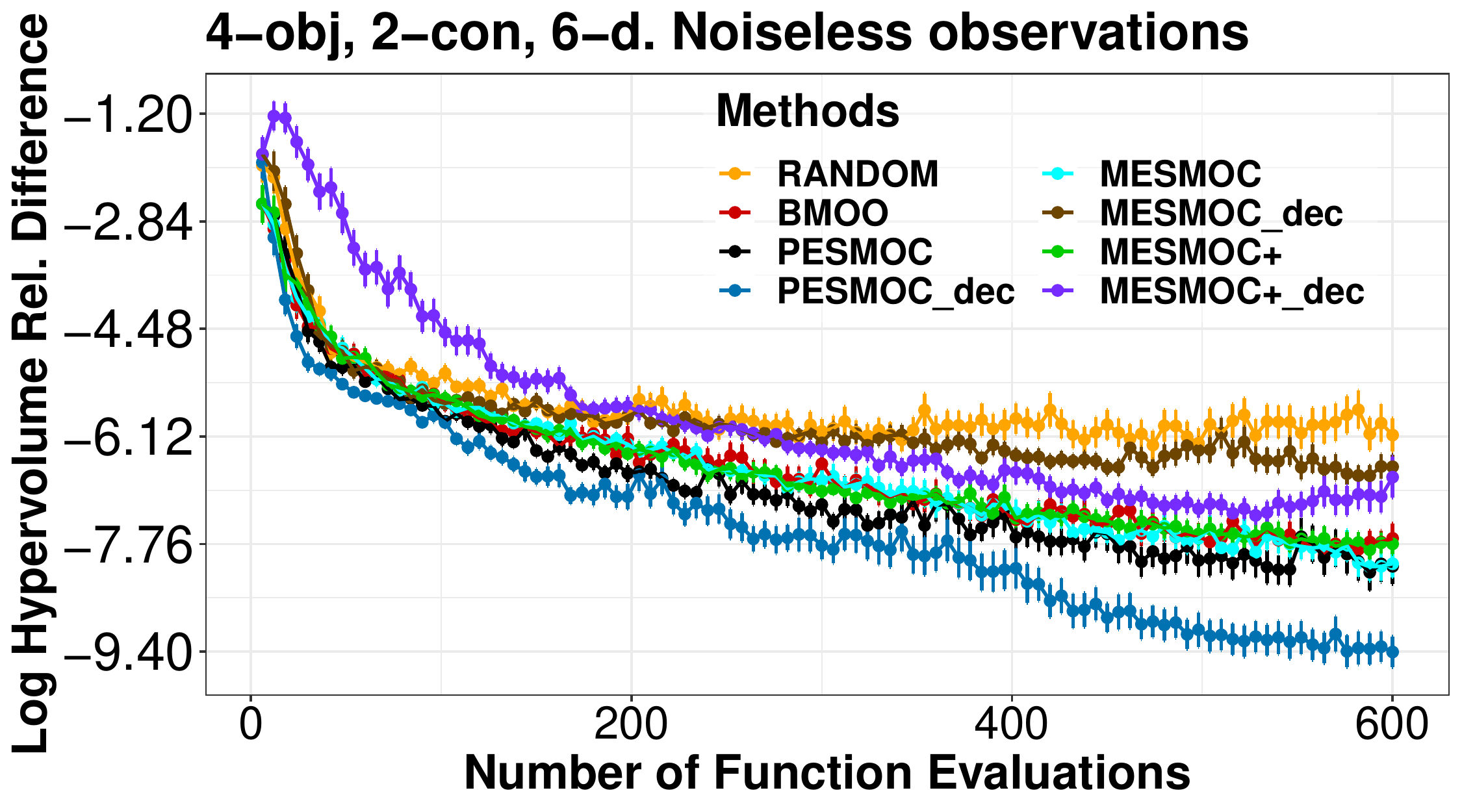} \\
		\includegraphics[width=0.5\textwidth]{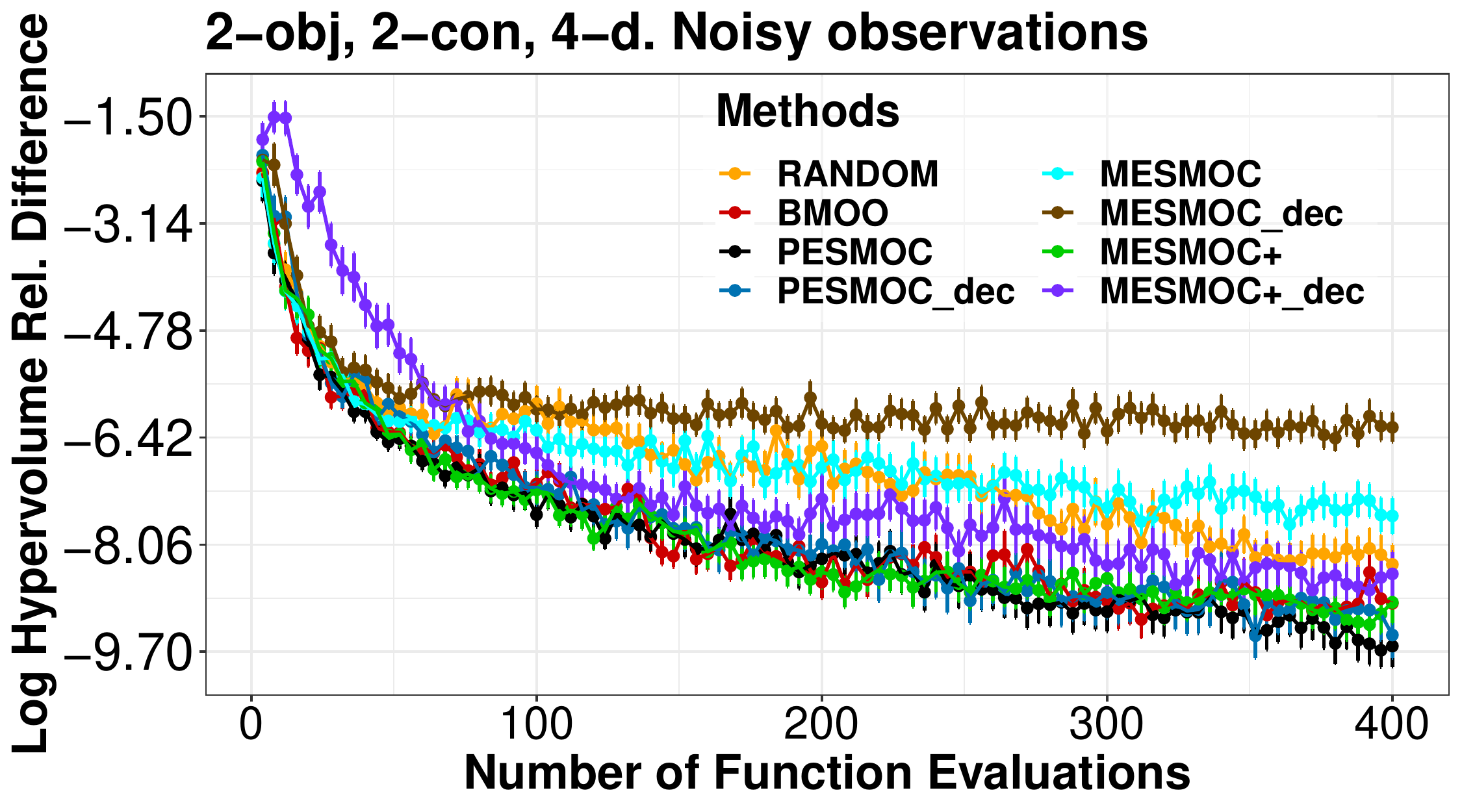}
		\includegraphics[width=0.5\textwidth]{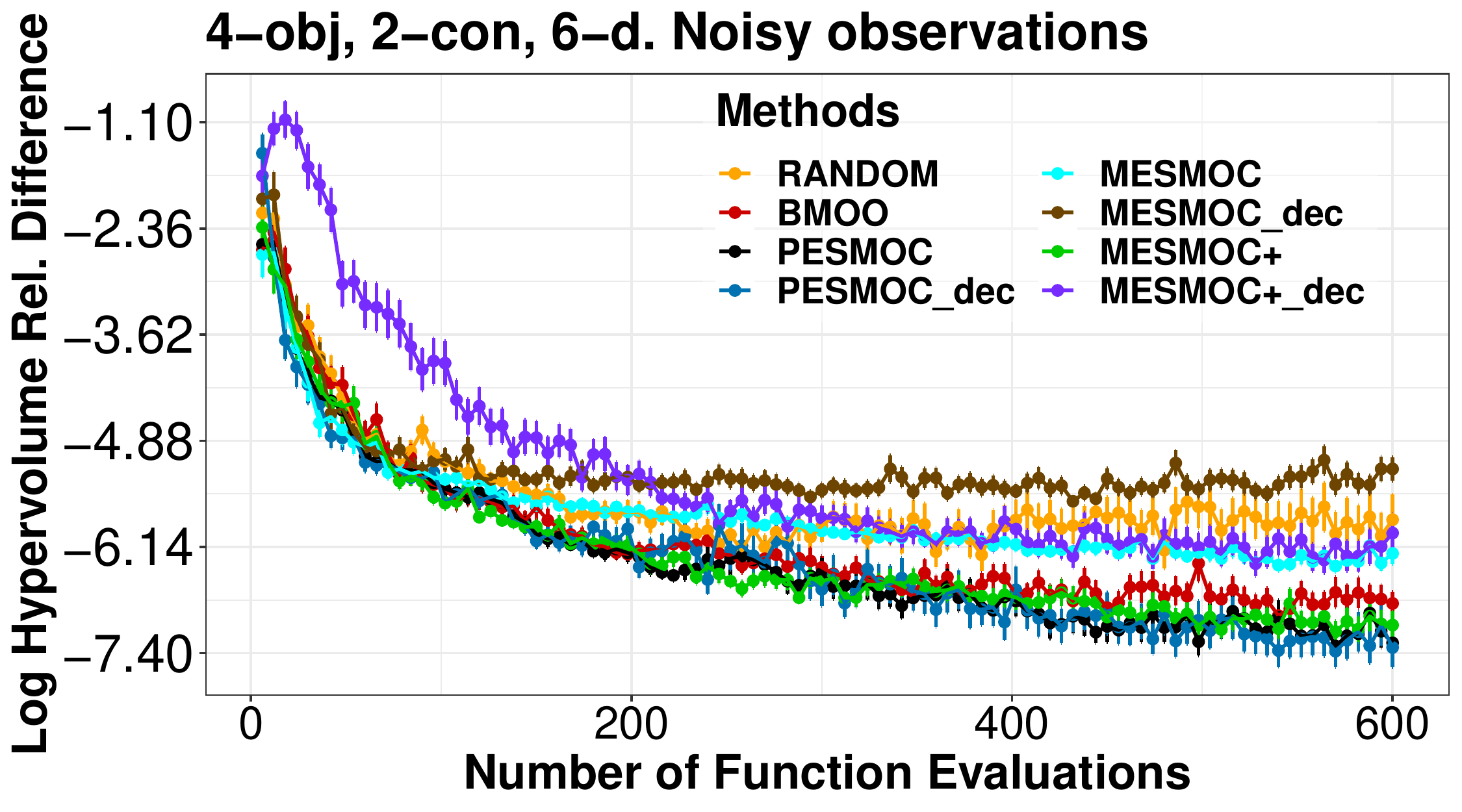}
	\end{tabular}}
	\caption{Avg. log hyper-volume relative difference between the recommendation of each 
	method at each iteration and the maximum hyper-volume in a 4-dimensional problem 
	(left-column) and in a 6-dimensional problem (right-column). We consider noiseless (top) and noisy 
	observations (bottom). Best seen in color. }
\end{figure*}

The results obtained by each method are shown in Fig. \ref{FIG:EXPSYNTHE}. 
We see that in the 4D experiment, in both scenarios, the best methods 
are MESMOC+, PESMOC and PESMOC$_\text{dec}$. MESMOC+$_\text{dec}$ also archives good results
when there is no noise. In these experiments MESMOC+ is superior to MESMOC, which 
performs poorly in the noisy settings. MESMOC$_\text{dec}$ also performs poorly in general. 
This is probably as a consequence of the poor approximation of the acquisition function in MESMOC 
and MESMOC$_\text{dec}$.  In the 6D experiments we observe similar results, 
but here MESMOC+$_\text{dec}$ gets significantly worse results than MESMOC+. This could be 
related to the removal of the logarithm in the acquisition function of MESMOC+$_\text{dec}$. 
In any case, in these experiments we observe that in general MESMOC+ and PESMOC give similar results, while
MESMOC seems to perform worse.

Table \ref{TB:TIMES} shows the average execution time in seconds per iteration 
of MESMOC+, MESMOC and PESMOC and their decoupled variants in the experiment 4D. 
We observe that the times of MESMOC+ and MESMOC+$_\text{dec}$ are significantly lower 
than those of PESMOC and PESMOC$_\text{dec}$, respectively.  This is because MESMOC+'s 
approximation is cheaper to compute. In particular, it reduces the entropy of the solution 
of the problem in the function space and uses ADF to approximate the conditional predictive 
distribution, instead of EP, like PESMOC. On the other hand, the total execution time of 
MESMOC is just a little lower than the one of MESMOC+, and although the runtime of 
MESMOC$_\text{dec}$ is half that MESMOC+$_\text{dec}$, its performance is much worse.

\begin{table}[ht]
	\caption{Avg. execution time per iteration (in sec.) in the 4D experiment. \strut} \label{TB:TIMES}
	\begin{center}
	\resizebox{\textwidth}{!}{\begin{tabular}{r@{$\pm$}l@{\hspace{3mm}}r@{$\pm$}l@{\hspace{3mm}}r@{$\pm$}l@{\hspace{3mm}}r@{$\pm$}l@{\hspace{3mm}}r@{$\pm$}l@{\hspace{3mm}}r@{$\pm$}l}
	\hline
		\multicolumn{2}{c}{\scriptsize \textbf{MESMOC+}}  &
		\multicolumn{2}{c}{\scriptsize \textbf{MESMOC+$_{\textrm{dec}}$}}  &
		\multicolumn{2}{c}{\scriptsize \textbf{MESMOC}}  &
		\multicolumn{2}{c}{\scriptsize \textbf{MESMOC$_{\textrm{dec}}$}}  &
		\multicolumn{2}{c}{\scriptsize \textbf{PESMOC}}  &
		\multicolumn{2}{c}{\scriptsize \textbf{PESMOC$_{\textrm{dec}}$}}  \\
		\hline 
		$12.48$ & $1.15$ &   
		$25.73$ & $3.94$ &   
		$10.34$ & $0.84$ &   
		$12.09$ & $0.83$ &   
		$29.71$ & $3.70$ &   
		$89.33$ & $5.36$ \\  
		\hline 
	\end{tabular}}
	\end{center}
\end{table}

\subsection{Finding an Optimal Ensemble}  \label{SB:ENSEMBLEEXP}
  
We tune the hyper-parameters of an ensemble of trees to classify the German dataset,
from the UCI repository \cite{dua2017}. This dataset has 1,000 instances, 20 attributes and 2 classes. 
The hyper-parameters of the ensemble are: the number of trees, the number of attributes to consider 
to split a node, the minimum number of samples to split a 
node, the probability of switching the class of each instance \cite{martinez2005switching} and 
the fraction of samples on which each tree is trained. We choose two 
objectives: to minimize the classification error, as estimated by a 10-fold-cv method, and to minimize the 
number of nodes of the trees in the ensemble. We also choose a constraint: the 
ensemble has to speed-up its average classification time by at least 25\% when using dynamic 
pruning \cite{hernandez2008statistical}. 

Both, the objectives and the constraints, can be evaluated separately as the total number 
of nodes is estimated by building only once the ensemble without leaving any data 
aside for validation. By contrast, the CV approach used to estimate the 
ensemble error requires to build several ensembles on subsets of the data. Similarly, 
evaluating the constraint involves building a lookup table. This table is expensive 
to build and is different for each ensemble size. 

\begin{figure*}[tbh]\label{FIG:EXPREAL}
	\centering
	\resizebox{\textwidth}{!}{\begin{tabular}{cc}
		\includegraphics[width=0.5\textwidth]{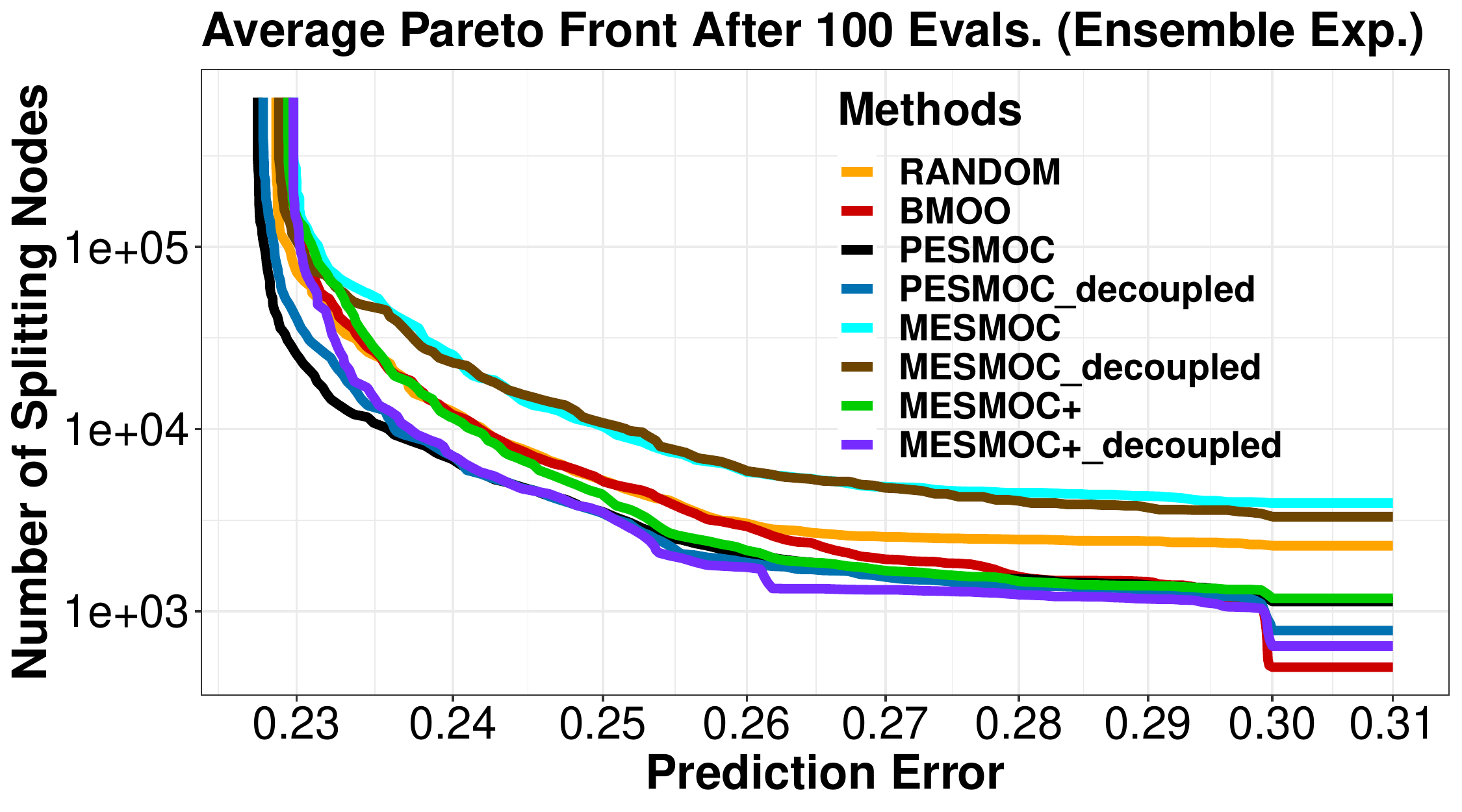}
		\includegraphics[width=0.5\textwidth]{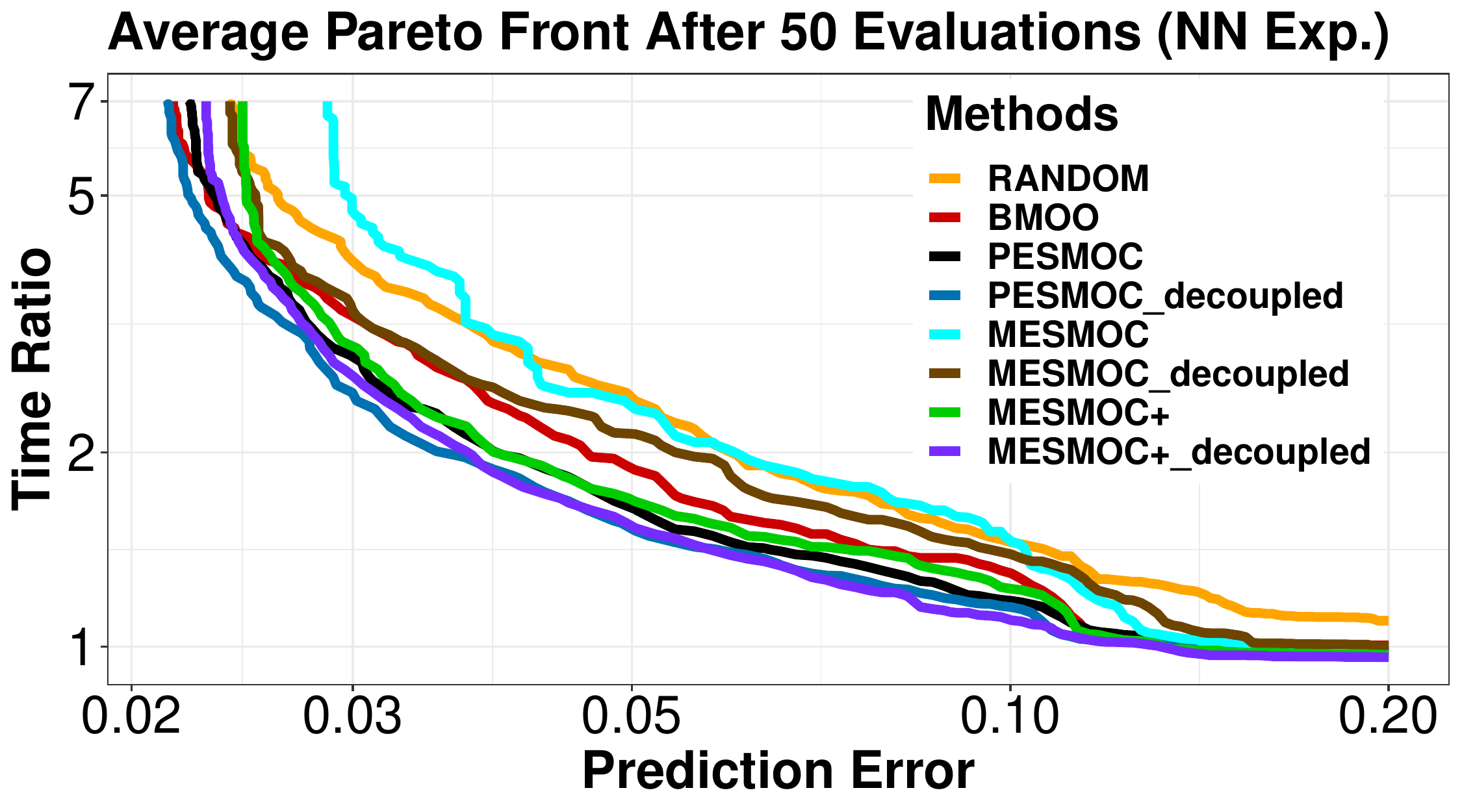} \\
		\includegraphics[width=0.5\textwidth]{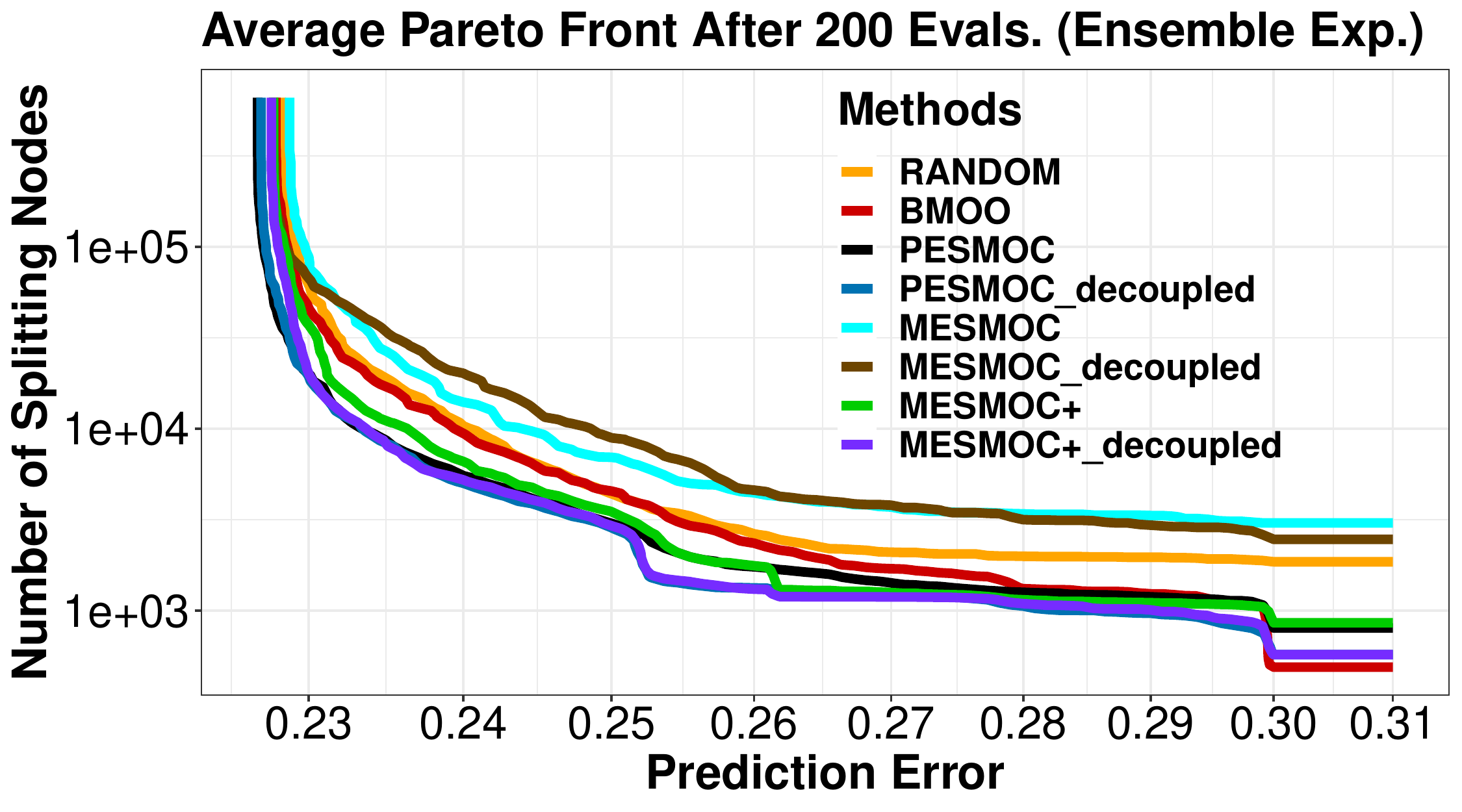}
		\includegraphics[width=0.5\textwidth]{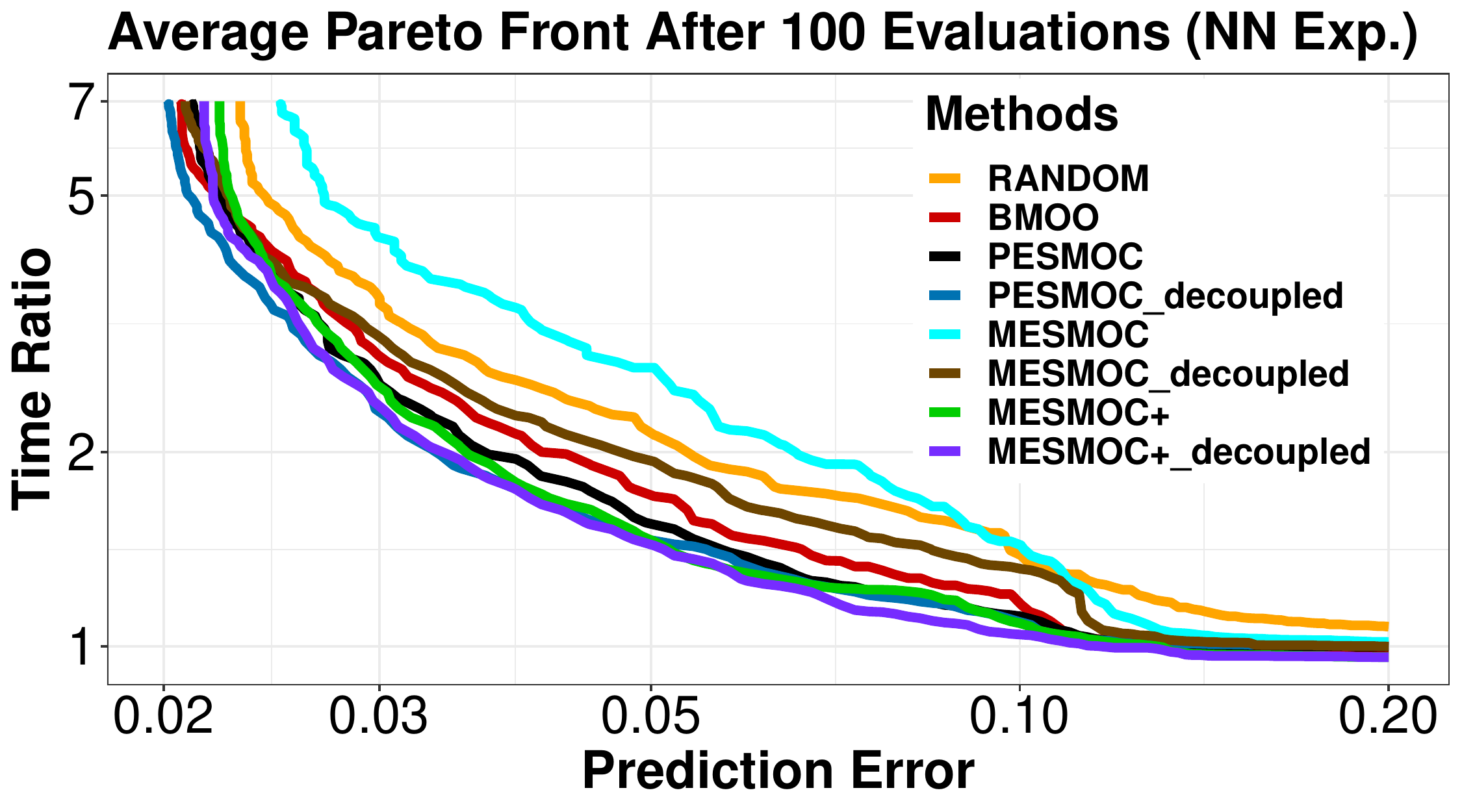}
	\end{tabular}}
	\caption{Avg. Pareto front of each method when finding an optimal ensemble (left-column) 
	and when finding an optimal neural network (right-column). Best seen in color. }
\end{figure*}

The left column of the Fig. \ref{FIG:EXPREAL} shows the average Pareto fronts 
obtained by each method after 100 and 200 evaluations. Since we are minimizing, 
the greater the area above the average Pareto front of a method, the better 
that method is, and the average hyper-volume of the solution is bigger.
Fig. \ref{FIG:EXPREAL} shows that MESMOC+$_\text{dec}$ and PESMOC$_\text{dec}$ obtain 
the best results. Although, MESMOC+$_\text{dec}$ 
obtains smaller ensembles but with more error, and PESMOC$_\text{dec}$ obtains larger 
ensembles but with less error. After 200 evaluations, the average Pareto 
front obtained by MESMOC+ has a similar quality to that of PESMOC, 
in terms of the hyper-volume. MESMOC+ finds better ensembles with an
error above 25\%. By contrast,  PESMOC finds better ensembles with a small error.
We can also see that MESMOC and MESMOC$_\text{dec}$ perform quite poorly. 
We believe this is a consequence of the difficulty of finding feasible solutions,
which means that these methods will significantly constrain the optimization of the acquisition
function, as described in Section \ref{SEC:RELATEDWORK}.
Table \ref{TB:ENSEMBLEHV} shows the average hyper-volume of the Pareto front 
found by each method. We observe see that the largest hyper-volume is obtained 
by MESMOC+$_\text{dec}$ and PESMOC$_\text{dec}$ after 100 and 200 evaluations, respectively, 
and their results are very similar.

\begin{table*}[t!]
	\caption{Average hyper-volume of each method after 100 and 200 evaluations in 
	the problem of finding an optimal ensemble. Similar results after 50 and 100 evaluations in the problem
	of finding an optimal neural network. The best result is bolded and the second best is underlined. \strut} \label{TB:ENSEMBLEHV}
	\begin{center}
			\begin{tabular}{lr@{$\pm$}lr@{$\pm$}lr@{$\pm$}lr@{$\pm$}lr@{$\pm$}l}
			\hline
    			\textbf{} 
    			& \multicolumn{4}{c}{\textbf{Ensemble}}
    			& \multicolumn{4}{c}{\textbf{Neural Network}} \\
    			\cmidrule(l){2-5}
    			\cmidrule(l){6-9}
    				\textbf{Method} &
    				\multicolumn{2}{c}{\textbf{100 Evals.}} &
    				\multicolumn{2}{c}{\textbf{200 Evals.}} &
    				\multicolumn{2}{c}{\textbf{50 Evals.}}  &
    				\multicolumn{2}{c}{\textbf{100 Evals.}} \\
    				\hline 
    				\text{MESMOC+}
    				& $0.293$ & $0.001$
    				& $0.322$ & $0.001$
    				& $47.84$ & $0.119$
    				& $53.90$ & $0.043$ \\
    				\text{MESMOC+$_{\textrm{dec}}$}
    				& $\boldsymbol{0.317}$ & $\boldsymbol{0.002}$
    				& $\underline{0.339}$ & $\underline{0.001}$
    				& $\underline{48.70}$ & $\underline{0.072}$
    				& $\underline{54.32}$ & $\underline{0.051}$ \\
    				\text{MESMOC}
    				& $0.220$ & $0.002$
    				& $0.243$ & $0.002$
    				& $45.24$ & $0.361$
    				& $46.27$ & $0.461$ \\
    				\text{MESMOC$_{\textrm{dec}}$}
    				& $0.215$ & $0.004$
    				& $0.234$ & $0.005$
    				& $45.70$ & $0.634$
    				& $49.65$ & $0.166$ \\
    				\text{PESMOC}
    				& $0.310$ & $0.001$
    				& $0.327$ & $0.001$
    				& $48.58$ & $0.074$
    				& $53.97$ & $0.057$ \\
    				\text{PESMOC$_{\textrm{dec}}$}
    				& $\underline{0.312}$ & $\underline{0.001}$
    				& $\boldsymbol{0.340}$ & $\boldsymbol{0.001}$
    				& $\boldsymbol{48.94}$ & $\boldsymbol{0.055}$
    				& $\boldsymbol{54.44}$ & $\boldsymbol{0.041}$ \\
    				\text{BMOO}
    				& $0.294$ & $0.001$
    				& $0.310$ & $0.001$
    				& $47.46$ & $0.261$
    				& $53.67$ & $0.085$ \\
    				\text{RANDOM}
    				& $0.264$ & $0.001$
    				& $0.280$ & $0.001$
    				& $46.23$ & $0.132$
    				& $51.99$ & $0.098$ \\
    				\hline
			\end{tabular}
	\end{center}
\end{table*}

\begin{figure*}[tbh]\label{FIG:REALEVALS}
	\centering
	\resizebox{\textwidth}{!}{\begin{tabular}{cc}
		\includegraphics[width=0.5\textwidth]{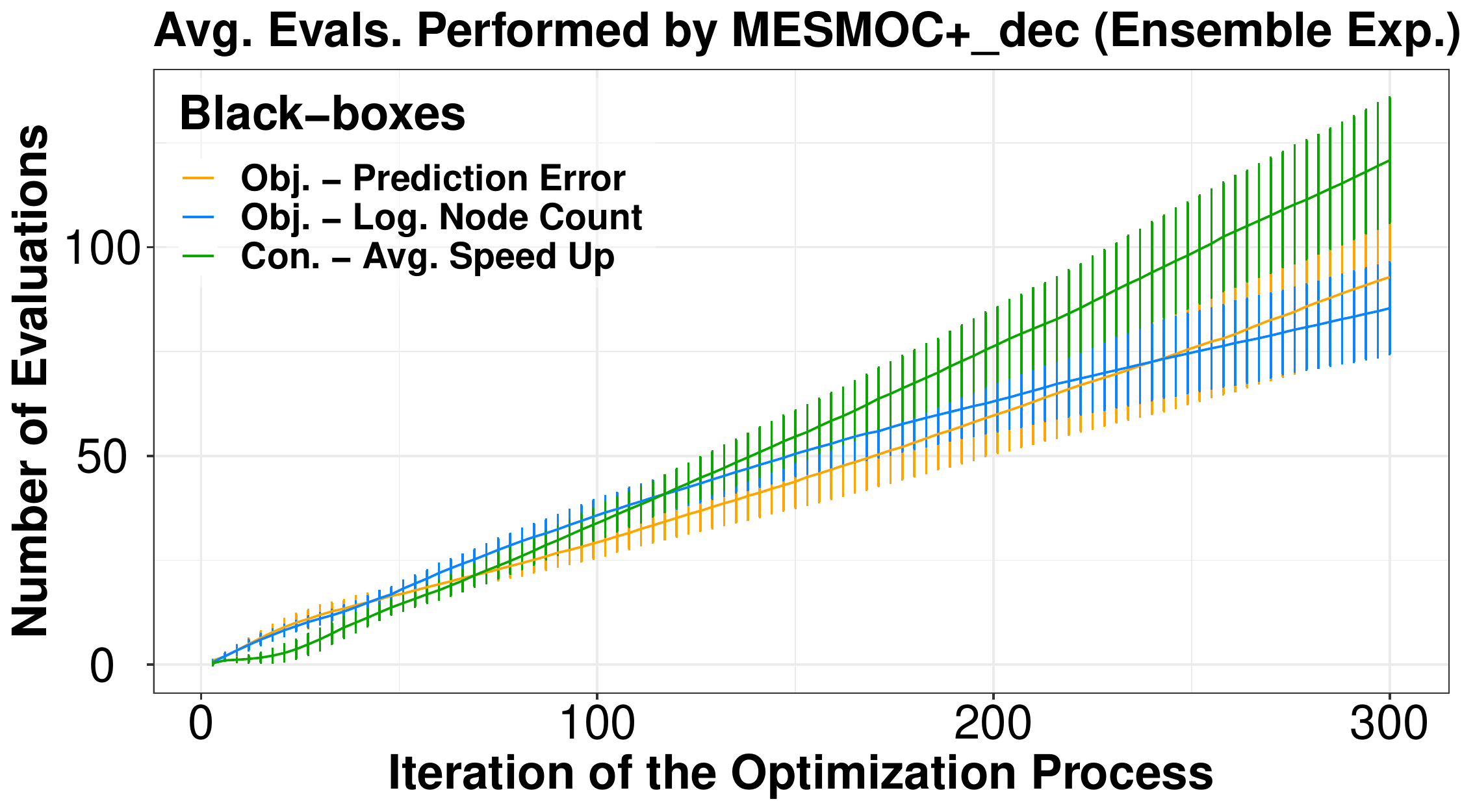}
		\includegraphics[width=0.5\textwidth]{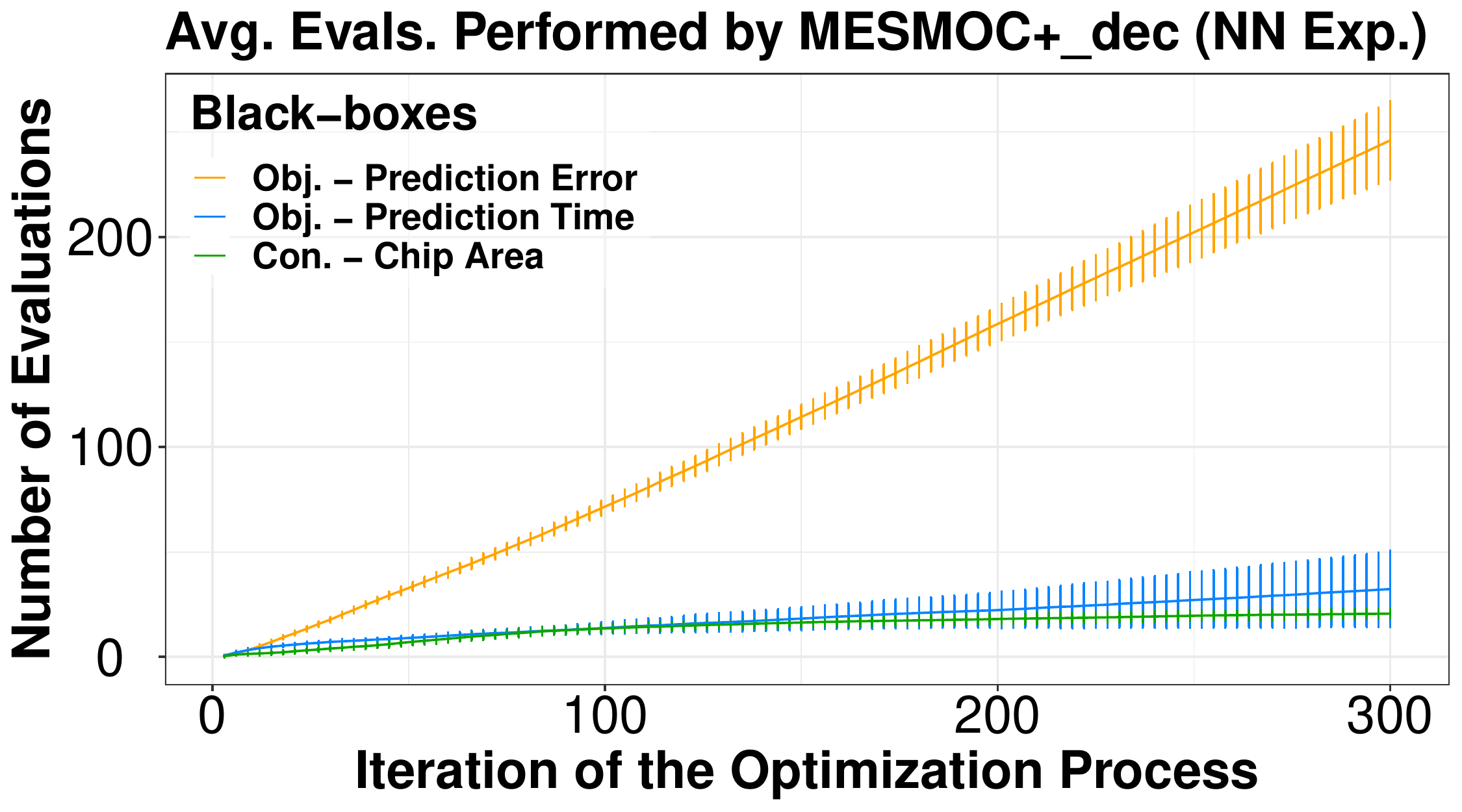}
	\end{tabular}}
	\caption{Number of evaluations performed by MESMOC+$_\text{dec}$ for each black-box in the problem
	of finding an optimal ensemble (left) and the problem of finding an optimal neural network (right). Best seen in color. }
\end{figure*}

Fig. \ref{FIG:REALEVALS} shows the number of evaluations of each black-box performed
by MESMOC+$_\text{dec}$. It evaluates approximately the same number of times each 
black-box. Therefore, the advantages of the decoupled setting come in this case 
from the fact that it can choose different input locations at which to evaluate 
each black-box, at each iteration. By contrast, the coupled version of MESMOC+ always 
evaluates all the black-boxes, at each iteration, on the same candidate point, which the
one maximizing the acquisition.

\subsection{Finding an Optimal Neural Network}  \label{SB:NETSEXP}

We aim to tune the hyper-parameters of a deep neural network. In this experiment, 
we consider the MNIST \cite{lecun2010mnist} dataset, which contains 60,000 images
of $28 \times 28$ pixels of hand-written digits. We build the network 
using Keras. For the training the networks we use ADAM with the default 
parameters \cite{kingma2014adam}. We have divided the dataset into 50,000 instances 
for training and 10,000 for validation. The hyper-parameters to adjust 
are: the number of hidden layers, the number of neurons in each layer, the learning rate, the 
dropout probability \cite{srivastava2014dropout}, the level of $\ell_1$ and $\ell_2$ regularization, and
two parameters related to the codification of the neural network in a chip: the memory partition 
and the loop unrolling factor. See \cite{garrido2019predictive} for more details.
  
The goal is to minimize the validation error and the prediction time of the network.
The constraint chose invalidate all networks that when codified into a chip result in an area greater than one square millimeter. 
The calculation of the area needed by each network is made using the hardware simulator Aladdin \cite{shao2014aladdin}.  
Again, these objectives and constraint can be evaluated independently. The prediction time is measured as the
ratio with respect to the prediction time of the fastest network (\emph{i.e.}, the smallest one).

In the right column of the Fig. \ref{FIG:EXPREAL} we shows the average Pareto front of each method after 50 and 100 evaluations. 
PESMOC$_\text{dec}$ is the method that obtains the Pareto front with the highest hyper-volume, 
followed by MESMOC+$_\text{dec}$ and PESMOC. We can also see that after 100 evaluations there is not much 
difference between MESMOC+$_\text{dec}$ and PESMOC$_\text{dec}$, and MESMOC+ and PESMOC. However, the decoupled variant obtains significantly better results than the coupled one but performing the same number of evaluations.
We can also see that the performance of MESMOC and MESMOC$_\text{dec}$ is worse or similar to that of RANDOM.
Table \ref{TB:ENSEMBLEHV} displays the average hyper-volume of the Pareto front 
of each method. The highest hyper-volume is obtained by PESMOC$_\text{dec}$, closely followed by MESMOC+$_\text{dec}$.

Finally, the number of evaluations performed by MESMOC+$_\text{dec}$ are displayed in the Fig. \ref{FIG:EXPREAL}. 
We observe that most of the evaluations have been carried out on the black-box corresponding to
the prediction error. It is hence expected that this black-box is more difficult to optimize and hence the proposed
approach, MESMOC+$_\text{dec}$ focuses more on it.

\section{Conclusions}   \label{SEC:CONCLUSIONS}
 
We have developed MESMOC+, a method for multi-objective Bayesian optimization with 
constraints. MESMOC+ selects the next point to evaluate as the one that is expected to reduce 
the most the entropy of the solution of the optimization problem in the function space. Namely, the Pareto frontier.
Since MESMOC+'s acquisition is expressed as a sum of acquisition functions, one per each different black-box, 
its computational cost is linear with respect to the number of black-boxes. Moreover,  it can be used in a decoupled 
evaluation setting. In our experiments we have observed that MESMOC+ is competitive with other methods from
the state-of-the-art for Bayesian optimization, but its cost per iteration is significantly smaller. 
The approximation of the acquisition function performed by MESMOC+ is also more accurate than that of existing methods.
Furthermore, a decoupled evaluation setting shows that MESMOC+ can not only choose where to evaluate next, 
but also which black-box to evaluate. Finally, we have observed that sometimes the decoupled variant of MESMOC+ 
achieves significantly better results than those of standard MESMOC+.

\subsubsection*{Acknowledgements}
The authors gratefully acknowledge the use of the facilities of Centro de Computaci\'on Cient\'ifica (CCC) at Universidad Aut\'onoma de Madrid. The authors also acknowledge financial support from Spanish Plan Nacional I+D+i, grants TIN2016-76406-P and from PID2019-106827GB-I00 / AEI / 10.13039/501100011033. Daniel Fer\'andez-S\'anchez also acknowledges the financial support from the Universidad Aut\'onoma de Madrid through Convocatoria de Ayudas para el fomento de la Investigaci\'on en Estudios de M\'aster-UAM 2019.

\bibliographystyle{abbrv}
\bibliography{mesmoc_paper}

\begin{thebibliography}{10}

\bibitem{ariizumi2014expensive}
R.~Ariizumi, M.~Tesch, H.~Choset, and F.~Matsuno.
\newblock Expensive multiobjective optimization for robotics with consideration
  of heteroscedastic noise.
\newblock In {\em IEEE International Conference on Intelligent Robots and
  Systems}, pages 2230--2235. IEEE, 2014.

\bibitem{belakaria2019max}
S.~Belakaria, A.~Deshwal, and J.~R. Doppa.
\newblock Max-value entropy search for multi-objective {B}ayesian optimization.
\newblock In {\em International Conference on Neural Information Processing
  Systems (NeurIPS)}, pages 7825--7835, 2019.

\bibitem{belakaria2020max}
S.~Belakaria, A.~Deshwal, and J.~R. Doppa.
\newblock Max-value entropy search for multi-objective bayesian optimization
  with constraints.
\newblock In {\em {N}eurIPS {W}orkshop on {M}achine {L}earning and the
  {P}hysical {S}ciences}, 2020.

\bibitem{boyen1998tractable}
X.~Boyen and D.~Koller.
\newblock Tractable inference for complex stochastic processes.
\newblock {\em International Conference on Uncertainty in Artificial
  Intelligence}, pages 33--42, 1998.

\bibitem{brochu2009tutorial}
E.~Brochu, V.~M. Cora, and N.~De~Freitas.
\newblock A tutorial on {B}ayesian optimization of expensive cost functions,
  with application to active user modeling and hierarchical reinforcement
  learning.
\newblock {\em Technical {R}eport TR-2009-023, {U}niversity of {B}ritish
  {C}olumbia}, 2009.

\bibitem{collette2004multiobjective}
Y.~Collette and P.~Siarry.
\newblock {\em Multiobjective optimization: principles and case studies}.
\newblock Springer Science \& Business Media, 2004.

\bibitem{dua2017}
D.~Dua and C.~Graff.
\newblock {UCI} machine learning repository, 2017.

\bibitem{feliot2017bayesian}
P.~Feliot, J.~Bect, and E.~Vazquez.
\newblock A {B}ayesian approach to constrained single-and multi-objective
  optimization.
\newblock {\em Journal of Global Optimization}, 67(1-2):97--133, 2017.

\bibitem{garrido2019predictive}
E.~C. Garrido-Merch{\'a}n and D.~Hern{\'a}ndez-Lobato.
\newblock Predictive entropy search for multi-objective {B}ayesian optimization
  with constraints.
\newblock {\em Neurocomputing}, 361:50--68, 2019.

\bibitem{hennig2012entropy}
P.~Hennig and C.~J. Schuler.
\newblock Entropy search for information-efficient global optimization.
\newblock {\em Journal of Machine Learning Research}, 13(6):1809--1837, 2012.

\bibitem{hernandez2016predictive}
D.~Hern{\'a}ndez-Lobato, J.~Hernandez-Lobato, A.~Shah, and R.~Adams.
\newblock Predictive entropy search for multi-objective bayesian optimization.
\newblock In {\em International Conference on Machine Learning}, pages
  1492--1501. PMLR, 2016.

\bibitem{hernandez2008statistical}
D.~Hern{\'a}ndez-Lobato, G.~Martinez-Munoz, and A.~Su{\'a}rez.
\newblock Statistical instance-based pruning in ensembles of independent
  classifiers.
\newblock {\em IEEE Transactions on Pattern Analysis and Machine Intelligence},
  31(2):364--369, 2008.

\bibitem{hernandez2015predictive}
J.~M. Hern{\'a}ndez-Lobato, M.~Gelbart, M.~Hoffman, R.~Adams, and
  Z.~Ghahramani.
\newblock Predictive entropy search for bayesian optimization with unknown
  constraints.
\newblock In {\em International conference on machine learning}, pages
  1699--1707. PMLR, 2015.

\bibitem{hernandez2014predictive}
J.~M. Hern{\'a}ndez-Lobato, M.~W. Hoffman, and Z.~Ghahramani.
\newblock Predictive entropy search for efficient global optimization of
  black-box functions.
\newblock {\em Advances in neural information processing systems}, pages
  918--926, 2014.

\bibitem{kingma2014adam}
D.~P. Kingma and J.~Ba.
\newblock {ADAM}: A method for stochastic optimization.
\newblock {\em International Conference on Learning Representations}, 2014.

\bibitem{lecun2010mnist}
Y.~LeCun, C.~Cortes, and C.~J.~C. Burges.
\newblock {MNIST} handwritten digit database.
\newblock {\em AT\&T Labs [Online]. Available: http://yann. lecun.
  com/exdb/mnist}, 2, 2010.

\bibitem{martinez2005switching}
G.~Mart{\'\i}nez-Mu{\~n}oz and A.~Su{\'a}rez.
\newblock Switching class labels to generate classification ensembles.
\newblock {\em Pattern Recognition}, 38(10):1483--1494, 2005.

\bibitem{minka2001expectation}
T.~P. Minka.
\newblock Expectation propagation for approximate bayesian inference.
\newblock In {\em Uncertainty in Artificial Intelligence}, volume~17, pages
  362--369, 2001.

\bibitem{perrone2019constrained}
V.~Perrone, I.~Shcherbatyi, R.~Jenatton, C.~Archambeau, and M.~Seeger.
\newblock Constrained {B}ayesian optimization with max-value entropy search.
\newblock {\em NeurIPS Workshop on Meta-Learning}, 2019.

\bibitem{rahimi2007random}
A.~Rahimi, B.~Recht, et~al.
\newblock Random features for large-scale kernel machines.
\newblock In {\em NIPS}, volume~3, pages 1177--1184. Citeseer, 2007.

\bibitem{rasmussen2006gaussian}
C.~E. Rasmussen and C.~K. Williams.
\newblock {\em Gaussian Processes for Machine Learning}.
\newblock MIT press, 2006.

\bibitem{shahriari2015taking}
B.~Shahriari, K.~Swersky, Z.~Wang, R.~P. Adams, and N.~De~Freitas.
\newblock Taking the human out of the loop: A review of {B}ayesian
  optimization.
\newblock {\em Proceedings of the IEEE}, 104(1):148--175, 2015.

\bibitem{shao2014aladdin}
Y.~S. Shao, B.~Reagen, G.-Y. Wei, and D.~Brooks.
\newblock Aladdin: A pre-rtl, power-performance accelerator simulator enabling
  large design space exploration of customized architectures.
\newblock In {\em 2014 ACM/IEEE 41st International Symposium on Computer
  Architecture (ISCA)}, pages 97--108. IEEE, 2014.

\bibitem{snoek2012practical}
J.~Snoek, H.~Larochelle, and R.~P. Adams.
\newblock Practical {B}ayesian optimization of machine learning algorithms.
\newblock {\em Advances in neural information processing systems}, pages
  2951--2959, 2012.

\bibitem{srivastava2014dropout}
N.~Srivastava, G.~Hinton, A.~Krizhevsky, I.~Sutskever, and R.~Salakhutdinov.
\newblock Dropout: a simple way to prevent neural networks from overfitting.
\newblock {\em The journal of machine learning research}, 15(1):1929--1958,
  2014.

\bibitem{suzuki2020multi}
S.~Suzuki, S.~Takeno, T.~Tamura, K.~Shitara, and M.~Karasuyama.
\newblock Multi-objective bayesian optimization using pareto-frontier entropy.
\newblock In {\em International Conference on Machine Learning}, pages
  9279--9288. PMLR, 2020.

\bibitem{villemonteix2009informational}
J.~Villemonteix, E.~Vazquez, and E.~Walter.
\newblock An informational approach to the global optimization of
  expensive-to-evaluate functions.
\newblock {\em Journal of Global Optimization}, 44(4):509--534, 2009.

\bibitem{wang2017max}
Z.~Wang and S.~Jegelka.
\newblock Max-value entropy search for efficient bayesian optimization.
\newblock In {\em International Conference on Machine Learning}, pages
  3627--3635. PMLR, 2017.

\bibitem{wilson2020efficiently}
J.~Wilson, V.~Borovitskiy, A.~Terenin, P.~Mostowsky, and M.~Deisenroth.
\newblock Efficiently sampling functions from gaussian process posteriors.
\newblock In {\em International Conference on Machine Learning}, pages
  10292--10302. PMLR, 2020.

\end{thebibliography}

\appendix

\section{Obtaining the Approximate Truncated Gaussians by ADF}

\subsection{Introduction to ADF}

  In this section, we will explain how Assumed Density Filtering (ADF) is used to approximate the predictive distributions conditioned 
  to the Pareto $\mathcal{Y}^\star$ front by truncated Gaussians. ADF is a technique that is often used to calculate approximate posteriors 
  \cite{boyen1998tractable}. When ADF is used to approximate a distribution of interest to $p(\mathbf{a})$, a distribution $q(\mathbf{a})$ 
  is first chosen from a family of distributions that it is convenient for us to use. This $q(\mathbf{a})$ distribution is adjusted to 
  approximate the target distribution $p(\mathbf{a})$. To adjust $q(\mathbf{a})$, ADF minimizes the Kullback-Leibler divergence between 
  $p(\mathbf{a})$ and $q(\mathbf{a})$, i.e. it minimizes $KL(p(\mathbf{a})||q(\mathbf{a}))$. We have chosen that $q(\mathbf{a})$ is a 
  truncated Gaussian, thus it belongs to the exponential family. Minimizing the divergence of Kullback-Leibler when we are approaching one distribution by another that belongs to the exponential family is equivalent to matching moments between the two distributions.  Namely, 
  we are going to adjust the means and variances of several truncated Gaussian distributions to approximate the predictive distributions 
  conditioned to $\mathcal{Y}^\star$. This adjustment of means and variances is made while processing the points of a $\mathcal{Y}^\star$ sample.


\subsection{ADF Update Equations}
  
  In this section, we will obtain the equations that allow us to update the means and variances. The expression of the predictive 
  distribution conditioned to $\mathcal{Y}^\star$ is given by:
\begin{equation} \label{EQ:CPDM1}
	p(\mathbf{f}, \mathbf{c}| \mathcal{D}, \mathbf{x}, \mathcal{Y}^\star)
	=
	Z^{-1}
	p(\mathbf{f}, \mathbf{c}| \mathcal{D}, \mathbf{x})
	p(\mathcal{Y}^\star| \mathbf{f}, \mathbf{c})
\end{equation}
  since this distribution can be expressed as:
\begin{equation}
	Z
	=
	\int t(\mathbf{a}) \mathcal{N}(\mathbf{a}|\boldsymbol{\mu}, \boldsymbol{\Sigma})d \mathbf{a},
	\qquad \qquad
	{p}(\mathbf{a})
	=
	Z^{-1} t(\mathbf{a}) \mathcal{N}(\mathbf{a}|\boldsymbol{\mu}, \boldsymbol{\Sigma})
	\notag
\end{equation}
  since the factor $\Omega (\mathbf{f}^\star, \mathbf{f}, \mathbf{c})$ of $p(\mathcal{Y}^\star| \mathbf{f}, \mathbf{c})$ is 
  $t(\mathbf{a})$ and the predictive distributions of the Gaussian processes are 
  $\mathcal{N}(\mathbf{a}|\boldsymbol{\mu}, \boldsymbol{\Sigma})$ where $\boldsymbol{\mu}$ y $\boldsymbol{\Sigma}$ are the vector 
  of means and the covariance matrix of a Gaussian, therefore we can use the following equations to iteratively adjust $q(\mathbf{a})$:
\begin{equation}\label{EQ:MEANVARIANCE}
	\begin{split}
		\mathbb{E}_{\hat{p}(\mathbf{a})}[\mathbf{a}]
		&=
		\boldsymbol{\mu} + \boldsymbol{\Sigma} \frac{\partial \log(Z)}{\partial \boldsymbol{\mu}} \\
		\mathbb{E}_{\hat{p}(\mathbf{a})}[\mathbf{a}\mathbf{a}^\textrm{T}]
		-
		\mathbb{E}_{\hat{p}(\mathbf{a})}[\mathbf{a}]
		\mathbb{E}_{\hat{p}(\mathbf{a})}[\mathbf{a}]^\textrm{T}
		&=
		\boldsymbol{\Sigma} - \boldsymbol{\Sigma} 
		\left(
		\frac{\partial \log(Z)}{\partial \boldsymbol{\mu}}
		\left(
		\frac{\partial \log(Z)}{\partial \boldsymbol{\mu}}
		\right)^\textrm{T}
		- 2 \frac{\partial \log(Z)}{\partial \boldsymbol{\Sigma}}
		\right) \boldsymbol{\Sigma} .
	\end{split}
\end{equation}

  
  Thus, to use ADF, we must calculate $Z$ and the partial derivatives of $\log(Z)$ with respect to the means and variances 
  of the objectives and constraints. The calculation of $Z$ is the following:

\begin{equation}
	\begin{split}
		Z
		&=
		\int
		p (\mathbf{f}, \mathbf{c} | \mathcal{D}, \mathbf{x}) \Omega (\mathbf{f}^\star, \mathbf{f}, \mathbf{c})
		d\mathbf{f} d\mathbf{c} \\
		&=
		\int
		p (\mathbf{f}, \mathbf{c} | \mathcal{D}, \mathbf{x})
		\left(
			1 - 
			\prod_{j=0}^C \Theta (c_j (\mathbf{x}))
			\prod_{k=0}^K \Theta \left(f_k^\star  - f_k (\mathbf{x}) \right)
		\right)
		d\mathbf{f} d\mathbf{c} \\
		&=
		1 -  \prod_{j=0}^C \Phi (\gamma^c_j) \prod_{k=0}^K \Phi (\gamma^f_k)
	\end{split}
\end{equation}
  where $\Phi (\cdot)$ is the cumulative probability distribution of Gaussian, and:

\begin{equation}
	\gamma^f_k = \frac{f_k^\star - m^f_k(\mathbf{x})}{(v^{f}_k(\mathbf{x}))^{1/2}}
	\qquad \qquad \qquad \qquad
	\gamma^c_j = \frac{m^c_j(\mathbf{x})}{(v^{c}_j(\mathbf{x}))^{1/2}}
\end{equation}
  where $m^f_k(\mathbf{x})$, $v^f_k(\mathbf{x})$, $m^c_k(\mathbf{x})$ and $v^c_k(\mathbf{x})$ are the mean and variance values 
  of the $k$-th and $j$-th predictive distributions of the objectives and constraints at $\mathbf{x}$. On the other hand, these 
  are the values of the derivatives 
  $\smash{\frac{\partial\log (Z)}{\partial m_k^f}}$, $\smash{\frac{\partial\log (Z)}{\partial v_k^f}}$, $\smash{\frac{\partial\log (Z)}{\partial m_j^c}}$ and $\smash{\frac{\partial\log (Z)}{\partial v_j^c}}$:
\begin{equation} \label{EQ:DERIVADASF}
		\frac{\partial\log (Z)}{\partial {m}^{f}_k}
		=
		\frac{(Z - 1)}{Z\Phi(\gamma^f_k)}
		\left(
			\frac{-\mathcal{N}(\gamma^f_k|0, 1)}{(v^{f}_k)^{1/2}}
		\right)
	\qquad
		\frac{\partial\log (Z)}{\partial v^{f}_k}
		=
		\frac{(Z - 1)}{Z\Phi(\gamma^f_k)}
		\left(
			\mathcal{N}(\gamma^f_k|0, 1) \frac{-\gamma^f_k}{2v^{f}_k}
		\right)
\end{equation}
\begin{equation} \label{EQ:DERIVADASC}
		\frac{\partial\log (Z)}{\partial {m}^{c}_j}
		=
		\frac{(Z - 1)}{Z\Phi(\gamma^c_j)}
		\left(
			\frac{\mathcal{N}(\gamma^c_j|0, 1)}{(v^{c}_j)^{1/2}}
		\right)
	\qquad
		\frac{\partial\log (Z)}{\partial v^{c}_j}
		=
		\frac{(Z - 1)}{Z\Phi(\gamma^c_j)}
		\left(
			\mathcal{N}(\gamma^f_j|0, 1) \frac{-\gamma^c_j}{2v^{c}_j}
		\right)
\end{equation}
where $v^f_k = v^f_k(\mathbf{x})$ and $v^{c}_j= v^{c}_j(\mathbf{x})$.

  We show the ADF algorithm in the Algorithm \ref{ALG:ADF}.  We can see that in each iteration the values of 
  $\mathbf{\tilde{m}}^{f}$, $\mathbf{\tilde{v}}^{f}$, $\mathbf{\tilde{m}}^{c}$ and $\mathbf{\tilde{v}}^{c}$ are updated using 
  the values calculated in the derivatives of \eqref{EQ:DERIVADASF} and \eqref{EQ:DERIVADASC}. We can also see that the order 
  of processing the $\mathbf{f}^\star$ points of the Pareto front sample will influence the result of the means and variances. 
  For this reason, we have established this order as random.
  
\begin{algorithm}[H]
	\SetKwInput{KwInput}{Input}
	\DontPrintSemicolon
	
		\KwInput{$\mathbf{m}^{f}$,
			$\mathbf{v}^{f}$,
			$\mathbf{m}^{c}$ and
			$\mathbf{v}^{c}$}
		
		Initialize: $\mathbf{\tilde{m}}^{f} = \mathbf{{m}}^{f}$,
					$\mathbf{\tilde{v}}^{f} = \mathbf{{v}}^{f}$,
					$\mathbf{\tilde{m}}^{c} = \mathbf{{m}}^{c}$ and
					$\mathbf{\tilde{v}}^{c} = \mathbf{{v}}^{c}$\;
		\For{${\mathbf{each}}$ $\mathbf{f}^\star$ ${\mathbf{in}}$ $\mathcal{Y}^\star$ }{
			$\boldsymbol{\gamma}^f
			=
			(\mathbf{f}^\star - \mathbf{m}^f) / \sqrt{\mathbf{v}^{f}}$\;
			$\boldsymbol{\gamma}^c
			=
			\mathbf{m}^c / \sqrt{\mathbf{v}^{c}}$\;
			$Z = \smash{1 -  \prod_{j=0}^C \Phi (\gamma^c_j) \prod_{k=0}^K \Phi (\gamma^f_k)}$\;
			$\mathbf{\tilde{m}}^{f} = \mathbf{\tilde{m}}^{f}
			+ \mathbf{\tilde{v}}^{f}
			\frac{\partial \log(Z)}{\partial \mathbf{\tilde{m}}^{f}}$\;
			$\mathbf{\tilde{v}}^{f} = \mathbf{\tilde{v}}^{f}
			- \mathbf{\tilde{v}}^{f}
			\left(
				\frac{\partial \log(Z)}{\partial \mathbf{\tilde{m}}^{f}}
				\left(
				\frac{\partial \log(Z)}{\partial \mathbf{\tilde{m}}^{f}}
				\right)^\textrm{T}
				- 2 \frac{\partial \log(Z)}{\partial \mathbf{\tilde{v}}^{f}}
			\right)$$\mathbf{\tilde{v}}^{f}$\;
			$\mathbf{\tilde{m}}^{c} = \mathbf{\tilde{m}}^{c}
			+ \mathbf{\tilde{v}}^{c}
			\frac{\partial \log(Z)}{\partial \mathbf{\tilde{m}}^{f}}$\;
			$\mathbf{\tilde{v}}^{c} = \mathbf{\tilde{v}}^{c}
			- \mathbf{\tilde{v}}^{c}
			\left(
				\frac{\partial \log(Z)}{\partial \mathbf{\tilde{m}}^{c}}
				\left(
				\frac{\partial \log(Z)}{\partial \mathbf{\tilde{m}}^{c}}
				\right)^\textrm{T}
				- 2 \frac{\partial \log(Z)}{\partial \mathbf{\tilde{v}}^{c}}
			\right)$$\mathbf{\tilde{v}}^{c}$\;
		}
		\KwRet $\mathbf{\tilde{m}}^{f}$,
		$\mathbf{\tilde{v}}^{f}$,
		$\mathbf{\tilde{m}}^{c}$ and
		$\mathbf{\tilde{v}}^{c}$;
\caption{ADF Algorithm} \label{ALG:ADF}
\end{algorithm}

\section{Quality of the Approximation of the Acquisition Function in a Decoupled Setting} \label{SEC:QUALITYAPPROX}

In this section, we compare in a simple problem the acquisition function of MESMOC+ and MESMOC with the exact acquisition function
described in Eq. \eqref{EQ:MESMOC+INI2}. The evaluations have been performed in a decoupled way. The problem considered has only
two objectives and one constraint. In this setting, quadrature methods are feasible to evaluate the entropy of
$p(\mathbf{y}| \mathcal{D}, \mathbf{x}, \mathcal{Y}^\star)$ at a much higher computational cost. They are expected to provide an
approximation that is almost equal to that of the exact acquisition

\begin{figure*}[tbh]
	\centering
	\begin{tabular}{cc}
		\includegraphics[width=0.5\textwidth]{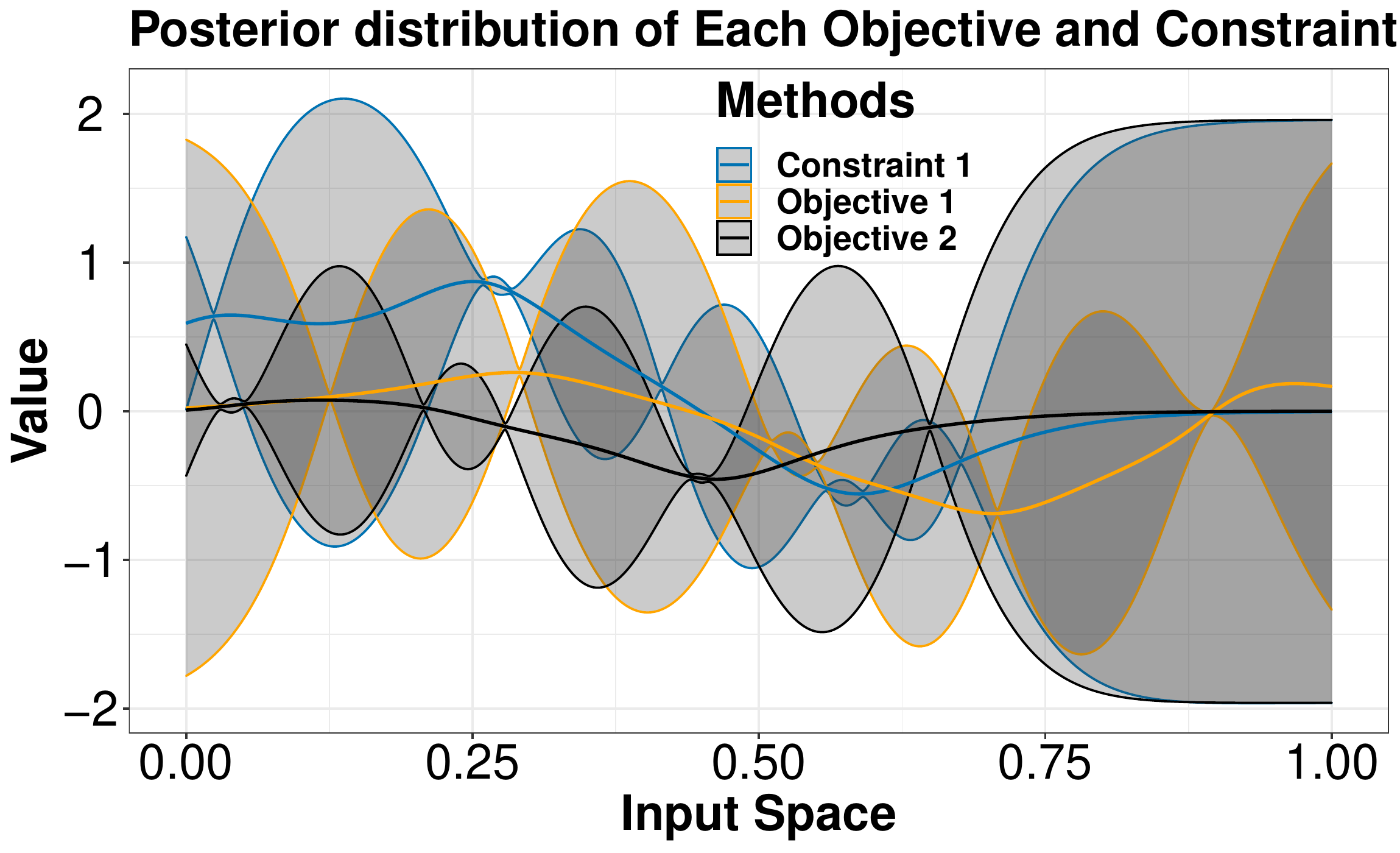}
		\includegraphics[width=0.5\textwidth]{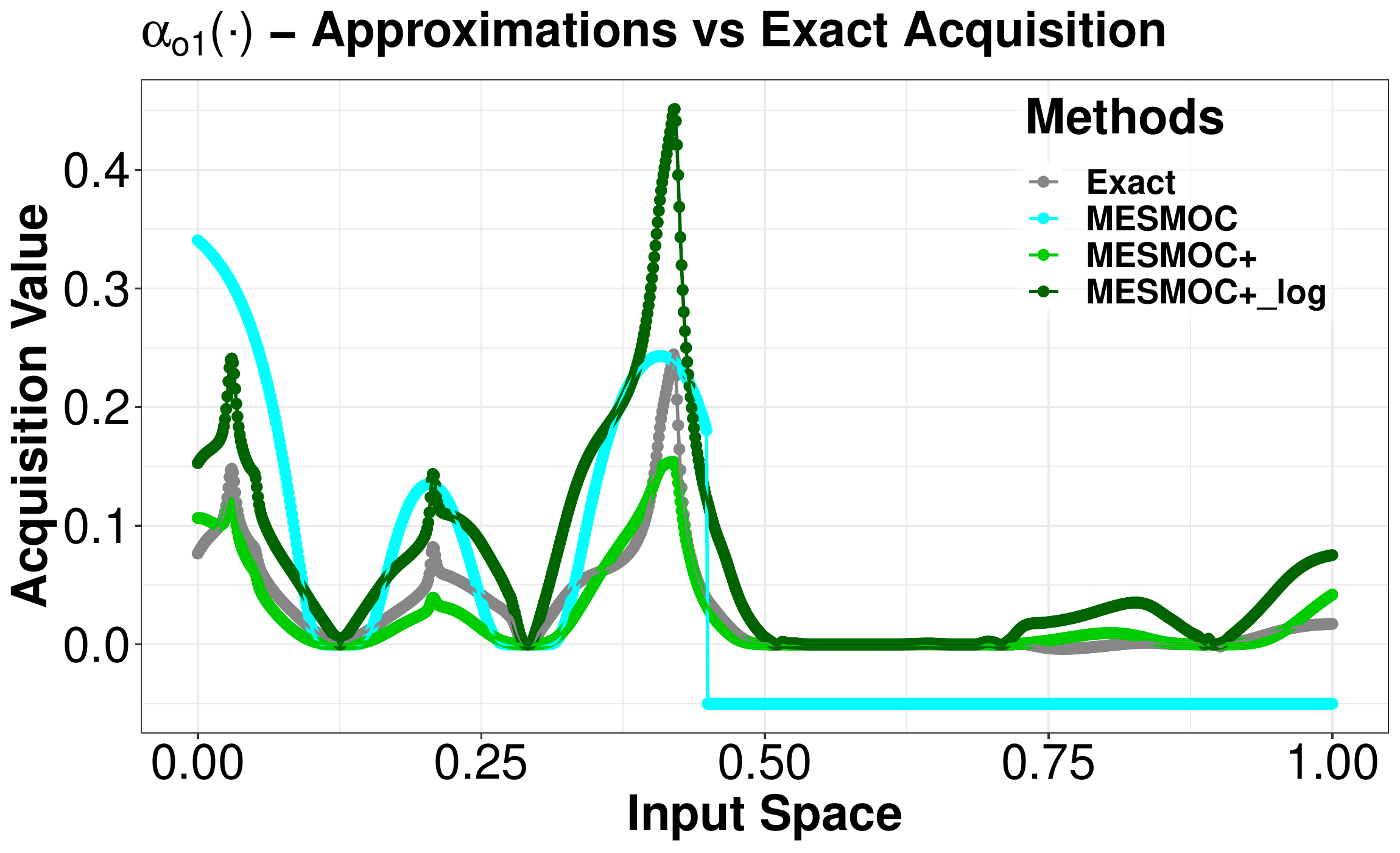} \\
		\includegraphics[width=0.5\textwidth]{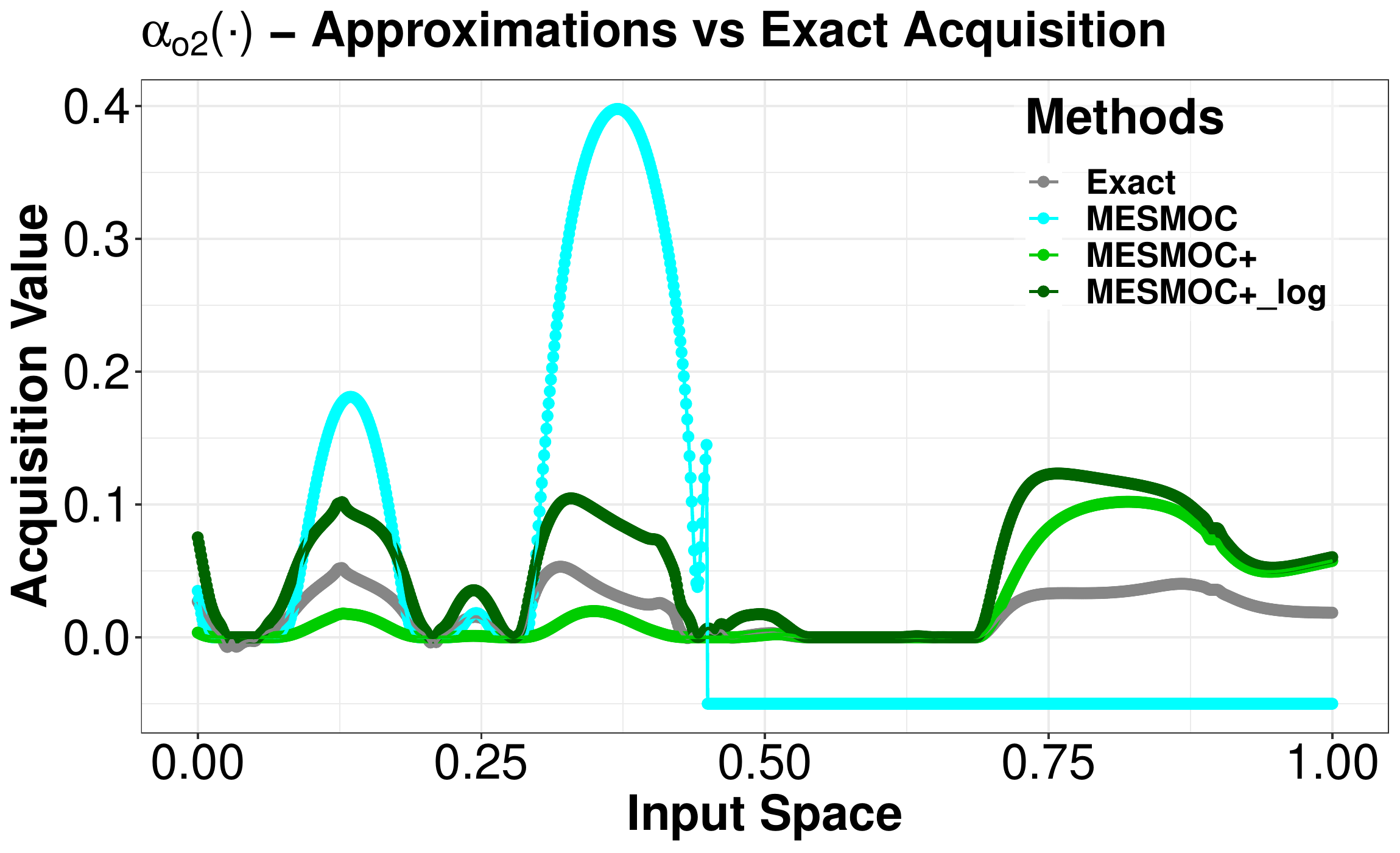}
		\includegraphics[width=0.5\textwidth]{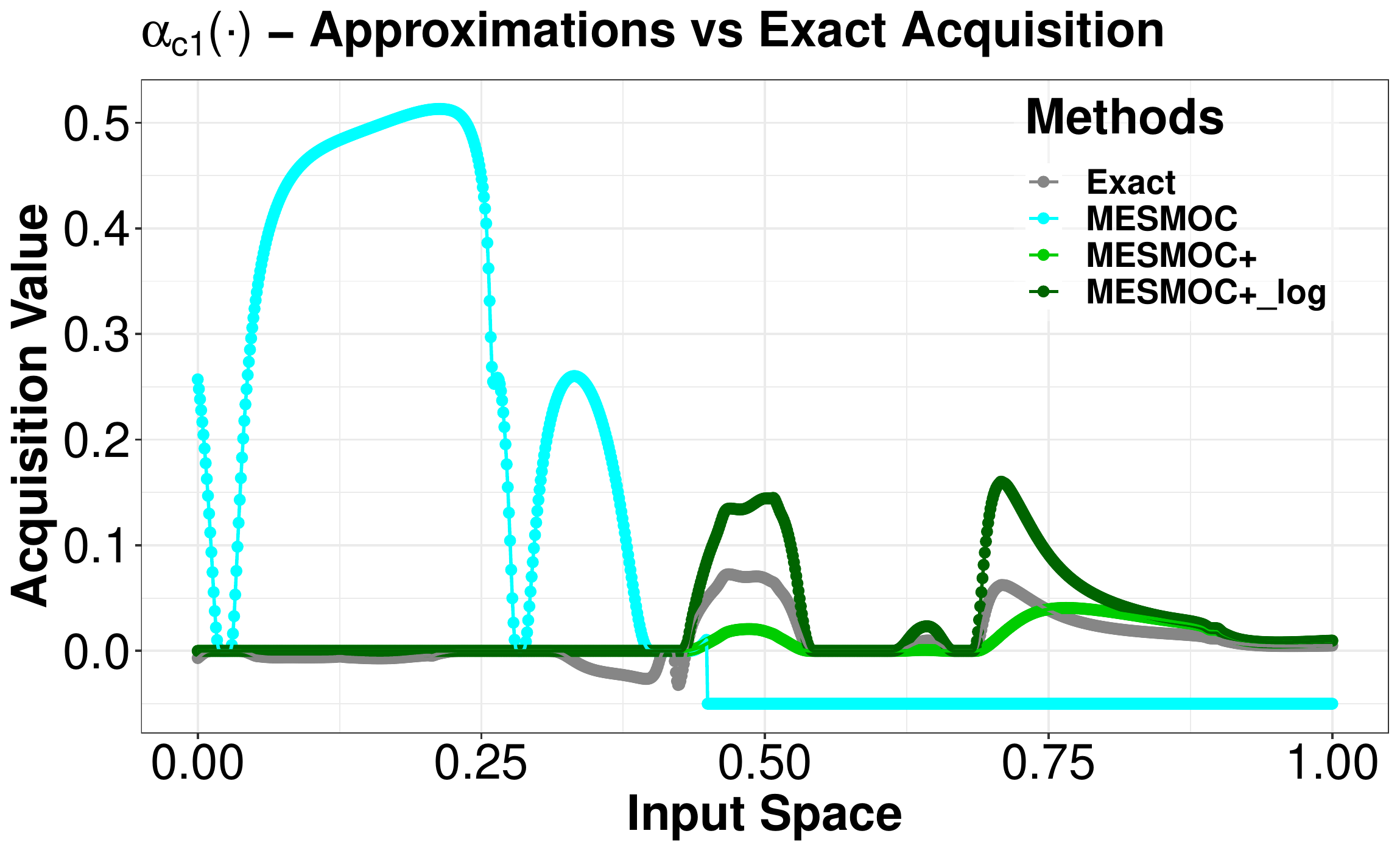}
	\end{tabular}
	\caption{(top-left) GP predictive distributions for the objectives and constraints. (bottom-left and right) The corresponding estimated  acquisition function of each method, MESMOC+ and MESMOC, and the exact acquisition (Exact) for each black-box. Best seen in color. } \label{FIG:CMPACQDEC}
\end{figure*}

The figure on the top-left of the Fig. \ref{FIG:CMPACQDEC} shows the current observations and predictive distributions for the objectives and constraints. 
The rest figures show the acquisition function for MESMOC and MESMOC+ for each black-box. In the case of MESMOC+, we show results for the 
proposed method and when the log of the variance is considered (MESMOC+log). See Eq. \eqref{EQ:MESMOC+FINAL}. Last, we also show the 
results of the quadrature method (Exact). We can observe that MESMOC+ and MESMOC+$_\text{log}$ are very similar to the exact in all the 
black-boxes since where MESMOC+ and MESMOC+$_\text{log}$ grow, the exact acquisition grows, and where the exact acquisition decreases, 
MESMOC+ and MESMOC+$_\text{log}$ also decrease. By contrast, the approximation of MESMOC does not look similar to the exact acquisition 
in any black-box. MESMOC avoids evaluations in the region where the GP mean of the constraint is negative. MESMOC’s acquisition there 
correspond to a constant value smaller than zero.

\end{document}